\documentclass[10pt,twocolumn,letterpaper]{article}

\usepackage[pagenumbers]{cvpr} 

\usepackage{graphicx}
\graphicspath{ {images/} }
\usepackage{amsmath}
\usepackage{amssymb}
\usepackage{mathtools}
\usepackage{bm}
\usepackage{bbm}
\usepackage{booktabs}
\usepackage{lipsum}
\usepackage{multirow}
\usepackage{tabularx}
\usepackage{makecell}
\usepackage{algorithm}
\usepackage{algpseudocode}
\usepackage{amssymb}
\usepackage{pifont}
\usepackage{adjustbox}
\usepackage[percent]{overpic}
\usepackage[title]{appendix}
\usepackage[accsupp]{axessibility}
\makeatletter
\@namedef{ver@everyshi.sty}{}
\makeatother
\usepackage{tikz}
\usepackage{pgfplots}
\usepackage{enumitem} 
\pgfplotsset{compat=1.5}

\usepackage{enumitem}
\setlist[itemize]{topsep={0pt},partopsep={0pt}}

\usepackage{caption}
\renewcommand{\arraystretch}{1.1}

\usepackage{xcolor}
\usepackage{colortbl}
\definecolor{gray}{gray}{0.95}

\usepackage[pagebackref,breaklinks,colorlinks]{hyperref}

\usepackage[capitalize]{cleveref}
\crefname{section}{Section.}{Secs.}
\Crefname{section}{Section}{Sections}
\Crefname{table}{Table}{Tables}
\crefname{table}{Tab.}{Tabs.}

\def\cvprPaperID{2977} 
\def\confName{CVPR}
\def\confYear{2023}

\begin{document}

\title{MetaPortrait: Identity-Preserving Talking Head Generation\\ with Fast Personalized Adaptation}
\vspace{-6cm}
\author{Bowen Zhang$^{1*}$ \qquad Chenyang Qi$^{2*}$ \qquad Pan Zhang$^{1}$ \qquad Bo Zhang$^{3}$\footnotemark[2] \qquad HsiangTao Wu$^{3}$\\ \qquad Dong Chen$^{2}$ \vspace{1pt} \qquad Qifeng Chen$^{2}$\footnotemark[2]
 \qquad Yong Wang$^{1}$ \qquad Fang Wen$^{3}$  \vspace{1pt}\\
$^{1}$USTC \qquad $^{2}$HKUST \qquad $^{3}$Microsoft
}
\twocolumn[{
\renewcommand\twocolumn[1][]{#1}
\maketitle
\vspace{-0.8cm}
\centering
\includegraphics[width=0.95\textwidth]{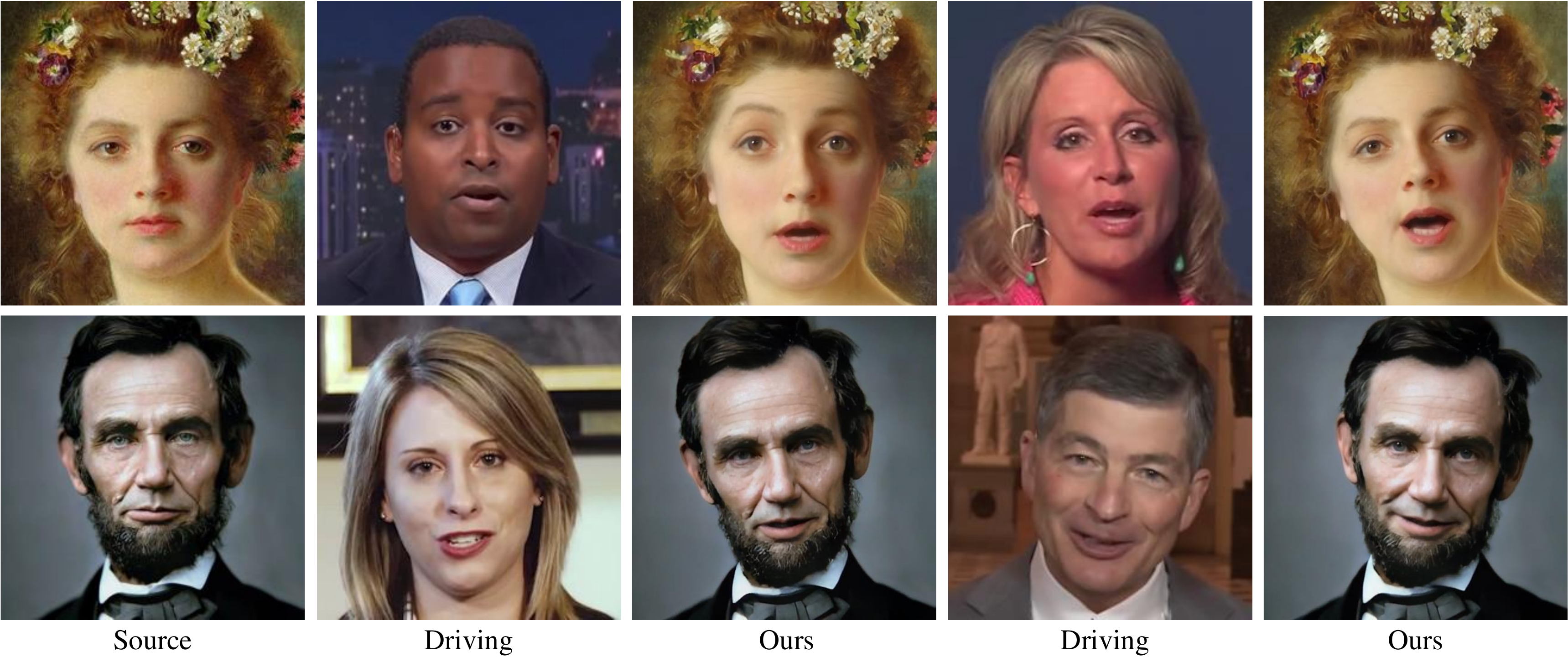}
\captionsetup{type=figure}
\caption{Our method yields identity-preserving talking head generation. See the \href{https://meta-portrait.github.io/}{webpage} for video demos.}
\label{fig:teaser}
\vspace{0.5cm}
}]

{
  \renewcommand{\thefootnote}%
    {\fnsymbol{footnote}}
  \footnotetext[1]{Equal contribution, interns at Microsoft Research.} \footnotetext[2]{Joint corresponding authors.}
}

\begin{abstract}
\vspace{-1em}
In this work, we propose an ID-preserving talking head generation framework, which advances previous methods in two aspects. First, as opposed to interpolating from sparse flow, we claim that dense landmarks are crucial to achieving accurate geometry-aware flow fields. Second, inspired by face-swapping methods, we adaptively fuse the source identity during synthesis, so that the network better preserves the key characteristics of the image portrait. Although the proposed model surpasses prior generation fidelity on established benchmarks, personalized fine-tuning is still needed to further make the talking head generation qualified for real usage. However, this process is rather computationally demanding that is unaffordable to standard users. To alleviate this, we propose a fast adaptation model using a meta-learning approach. The learned model can be adapted to a high-quality personalized model as fast as 30 seconds. 
Last but not least, a spatial-temporal enhancement module is proposed to improve the fine details while ensuring temporal coherency. Extensive experiments prove the significant superiority of our approach over the state of the arts in both one-shot and personalized settings.
\end{abstract}

\vspace{-2em}
\section{Introduction}
\label{sec:intro}

Talking head generation~\cite{YuruiRen2021PIRendererCP,AliaksandrSiarohin2019FirstOM,mallya2022implicit,Drobyshev22MegaPortraits,FeiYin2022StyleHEATOH,wiles2018x2face,ba2016layer,doukas2020headgan,TingChunWang2020OneShotFN,ouyang2022real} has found extensive applications in face-to-face live chat, virtual reality and virtual avatars in games and videos. In this paper, we aim to synthesize a realistic talking head with a single source image (one-shot) that provides the appearance of a given person while being animatable according to the motion of the driving person. Recently, considerable progress has been made with neural rendering techniques, bypassing the sophisticated 3D human modeling process and expensive driving sensors. While these works attain increasing fidelity and higher rendering resolution, identity preserving remains a challenging issue since the human vision system is particularly sensitive to any nuanced deviation from the person's facial geometry.

Prior arts mainly focus on learning a geometry-aware warping field, either by interpolating from sparse 2D/3D landmarks or leveraging 3D face prior, \eg, 3D morphable face model (3DMM)~\cite{JamesBooth2016A3DMM,VolkerBlanz1999A3DMM}. However, fine-grained facial geometry may not be well described by a set of sparse landmarks or inaccurate face reconstruction. Indeed, the warping field, trained in a self-supervised manner rather than using accurate flow ground truth, can only model coarse geometry deformation, lacking the expressivity that captures the subtle semantic characteristics of the portrait. 

In this paper, we propose to better preserve the portrait identity in two ways. First, we claim that dense facial landmarks are sufficient for an accurate warping field prediction without the need for local affine transformation.
Specifically, we adopt a landmark prediction model~\cite{wood2022dense} trained on synthetic data~\cite{ErrollWood2021FakeIT}, yielding 669 head landmarks that offer significantly richer information on facial geometry. In addition, we build upon the face-swapping approach~\cite{LingzhiLi2019FaceShifterTH} and propose to enhance the perceptual identity by attentionally fusing the identity feature of the source portrait while retaining the pose and expression of the intermediate warping. Equipped with these two improvements, our one-shot model demonstrates a significant advantage over prior works in terms of both image quality and perceptual identity preservation when animating in-the-wild portraits.

While our one-shot talking head model has achieved state-of-the-art quality, it is still infeasible to guarantee satisfactory synthesis results because such a one-shot setting is inherently ill-posed---one may never hallucinate the person-specific facial shape and occluded content from a single photo. Hence, ultimately we encounter the \emph{uncanny valley}~\cite{shin2019uncanny} that a user becomes uncomfortable as the synthesis results approach to realism. To circumvent this, 
one workaround is to finetune the model using several minutes of a personal video. Such personalized training has been widely adopted in industry to ensure product-level quality, yet this process is computationally expensive, which greatly limits its use scenarios. Thus, speeding up this \emph{personalized training}, a task previously under-explored, is of great significance to the application of talking head synthesis.

We propose to achieve fast personalization with meta-learning. The key idea is to find an initialization model that can be easily adapted to a given identity with limited training iterations. To this end, we resort to a meta-learning approach~\cite{ChelseaFinn2017ModelAgnosticMF,AlexNichol2018ReptileAS} that finds success in quickly learning discriminative tasks, yet is rarely explored in generative tasks. Specifically, we optimize the model for specific personal data with a few iterations. In this way, we get a slightly fine-tuned personal model towards which we move the initialization model weight a little bit. Such meta-learned initialization allows us to train a personal model within 30 seconds, which is 3 times faster than a vanilla pretrained model while requiring less amount of personal data.

Moreover, we propose a novel temporal super-resolution network to enhance the resolution of the generated talking head video.
To do this, we leverage the generative prior to boost the high-frequency details for portraits and meanwhile take into account adjacent frames that are helpful to reduce temporal flickering. Finally, we reach temporally coherent video results of $512\times 512$ resolution with compelling facial details. In summary, this work innovates in the following aspects:
\begin{itemize}[leftmargin=*]
    \itemsep=-1.5mm
    \item We propose a carefully designed framework to significantly improve the identity-preserving capability when animating a one-shot in-the-wild portrait.
    \item To the best of our knowledge, we are the first to explore meta-learning to accelerate personalized training, thus obtaining ultra-high-quality results at affordable cost. 
    \item Our novel video super-resolution model effectively enhances details without introducing temporal flickering.
\end{itemize}

\section{Related Work}
\label{sec:related_work}

\noindent\textbf{2D-based talking head synthesis.}
Methods along this line~\cite{Siarohin2019MonkeyNet, AliaksandrSiarohin2019FirstOM,TingChunWang2020OneShotFN,mallya2022implicit,pang2023dpe} predict explicit warping flow by interpolating the sparse flow defined by 2D landmarks. 
FOMM~\cite{AliaksandrSiarohin2019FirstOM} assumes local affine transformation for flow interpolation.
Recently, Mallya ~\etal~\cite{mallya2022implicit} computes landmarks from multiple source images using an attention mechanism. However, the landmarks learned in an unsupervised manner are too sparse (\eg, 20 in Face-Vid2Vid~\cite{TingChunWang2020OneShotFN}) to be interpolated into dense flows. Using predefined facial landmarks~\cite{XiaojieGuo2019PFLDAP} is also a straightforward approach~\cite{tripathy2021facegan,ha2020marionette,EgorZakharov2020FastBN,RuiqiZhao2021SparseTD} to represent the motion of driving images. For example, PFLD~\cite{XiaojieGuo2019PFLDAP} uses 98 facial landmarks. However, they could not generate an accurate warping flow since the landmarks are not dense enough.

\noindent\textbf{Talking head synthesis with 3D face prior.} 3D Morphable Models~\cite{JamesBooth2016A3DMM,VolkerBlanz1999A3DMM} represent a face image as PCA coefficients relating to identity, expression, and pose, which provides an easy tool to edit and render portrait images~\cite{HyeongwooKim2018DeepVP,kim2019neural,Geng2018WarpguidedGF,Fried2019Text-based-Editing,xing2023codetalker,tang2022explicitly}.
Some attempts~\cite{HyeongwooKim2018DeepVP,kim2019neural} render the animated faces by combining the identity coefficients of the source image and the motion coefficients of the driving video. Recent works~\cite{YuruiRen2021PIRendererCP,doukas2020headgan} predict a warping flow field for talking head synthesis using 3DMM. Although 3DMM-based methods allow explicit face control, using these coefficients to represent detailed face geometry and expression remains challenging. 

\begin{figure*}[ht]
    \centering
        \includegraphics[width=2.0\columnwidth]{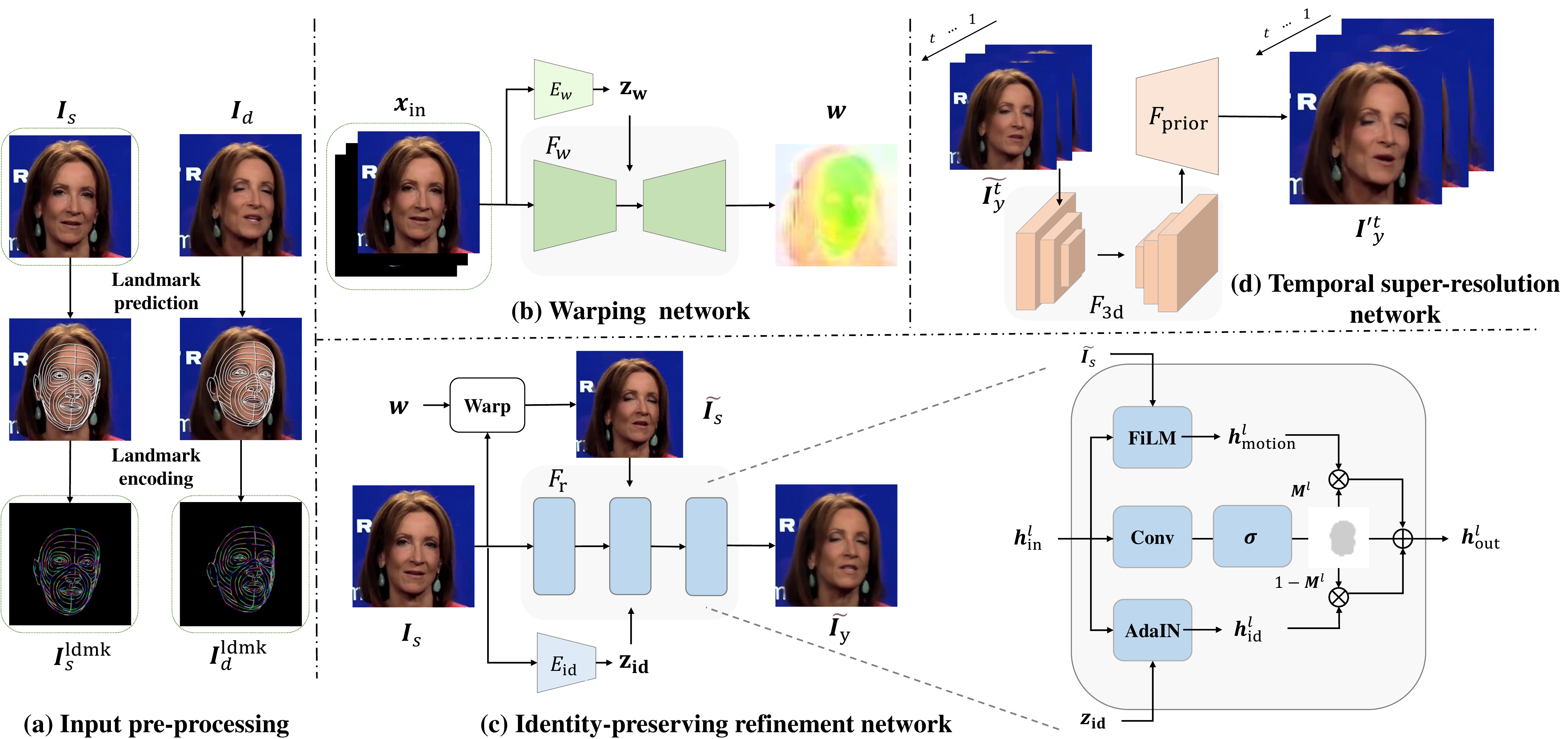}
    \caption{Overview of our one-shot framework. (a) Given a source image ${\bm{I}}_s$ and a driving video \{${\bm{I}}_d^1$, ${\bm{I}}_d^2$, $\cdots$, ${\bm{I}}_d^t$\}, we first extract their dense landmarks $( {\bm{I}}_s^{\textnormal{ldmk}}, \bm{I}_{\textnormal{d}}^{\textnormal{ldmk}})$ using a pretrained landmark detector. (b) Then, we estimate warping flows $\bm{w}$ between the source image and each driving frame according to concatenated input $\bm{x}_{\textnormal{in}}$. (c) We further refine the warped source input $\tilde{{\bm{I}}}_s$ using an ID-preserving network. (d) Finally, we enhance and upsample the $256\times 256$ results \{$\tilde{{\bm{I}}}_y^1$, $\tilde{{\bm{I}}}_y^2$, $\cdots$, $\tilde{{\bm{I}}}_y^t$\} to high-fidelity output \{${{\bm{I}}'}_y^1$, ${{\bm{I}}'}_y^2$, $\cdots$, ${{\bm{I}}'}_y^t$\} in $512\times 512$.} 
    \vspace{-0.5cm}
    \label{fig:overall_framework}
\end{figure*}

\noindent\textbf{Towards higher-resolution talking head.}
The video resolution of most talking head methods~\cite{YuruiRen2021PIRendererCP,AliaksandrSiarohin2019FirstOM,ba2016layer,wiles2018x2face} is $256\times256$, bounded by the available video datasets~\cite{JoonSonChung2018VoxCeleb2DS,ArshaNagrani2017VoxCelebAL}.
To enhance the visual quality of output videos, previous literature trains an additional single-image super-resolution network~\cite{Drobyshev22MegaPortraits,LingboYang2020HiFaceGANFR,wang2021gfpgan} or utilizes pretrained 2D StyleGAN~\cite{FeiYin2022StyleHEATOH,wang2021gfpgan}.
However, these methods upsample the original video in a frame-by-frame manner and ignore the temporal consistency of adjacent video frames. In this work, we propose a novel temporal super-resolution module to boost temporal consistency while preserving per-frame quality.

\noindent\textbf{Meta learning and fast adaptation.} The goal of meta-learning methods~\cite{ChelseaFinn2017ModelAgnosticMF, AlexNichol2018ReptileAS,EgorZakharov2019FewShotAL,Wang2021MetaAvatar} is to achieve good performance with a limited amount of training data. In this paper, we aim to quickly build a personalized model given a new identity, which correlates with the goal of meta-learning. Zakharov~\etal~\cite{EgorZakharov2019FewShotAL} propose to improve GAN inversion using a hyper-network. We leverage the idea of model-agnostic meta-learning (MAML)~\cite{ChelseaFinn2017ModelAgnosticMF} to obtain the best initialization that can be easily adapted to different persons.

\section{Method}
\label{sec:method}

Figure~\ref{fig:overall_framework} illustrates an overview of our synthesis framework. Given an image $\bm{I}_s$ of a person which we refer as source image and a sequence of $t$ driving video frames \{$\bm{I}_d^1$, $\bm{I}_d^2$, $\cdots$, $\bm{I}_d^t$\}, we aim to generate an output video \{${{\bm{I}}'}_y^1$, ${{\bm{I}}'}_y^2$, $\cdots$, ${{\bm{I}}'}_y^t$\} with the motions derived from the driving video while maintaining the identity of the source image. Section~\ref{sec:general_framework} introduces our one-shot model (Figure~\ref{fig:overall_framework}(a, b, c)) for identity-preserving talking head generation $\tilde{{\bm{I}}}_y$ at $256\times256$ resolution. We describe our meta-learning scheme in Section~\ref{sec:meta}, which allows fast adaption using a few images. In Section~\ref{sec:sr}, we propose a spatial-temporal enhancement network that generally improves the perceptual quality for both the one-shot and personalized models, yielding video frames with $512\times512$ resolution, as shown in Figure~\ref{fig:overall_framework}(d). 

\subsection{ID-preserving One-shot Base Model}
\label{sec:general_framework}

In this section, we will introduce our warping network using dense facial landmarks and an identity-aware refinement network for one-shot talking head synthesis.

\noindent\textbf{Warping prediction with dense landmarks.} 
To predict an accurate geometry-aware warping field, we claim that dense landmark prediction~\cite{wood2022dense} is the key to a geometry-aware warping field estimation. While dense facial landmarks are tedious to annotate, our dense landmark prediction is trained on synthetic faces~\cite{ErrollWood2021FakeIT}, which reliably produces 669 points covering the entire head, including the ears, eyeballs, and teeth, for in-the-wild faces. These dense landmarks capture rich information about the person's facial geometry and considerably ease the flow field prediction.

However, it is non-trivial to fully make use of these dense landmarks. A naive approach is to channel-wise concatenate the landmarks before feeding into the network, as previous works~\cite{AliaksandrSiarohin2019FirstOM,Siarohin2019MonkeyNet,TingChunWang2020OneShotFN}. However, processing such input is computationally demanding due to the inordinate number of input channels. Hence, we propose an efficient way to digest these landmarks. Specifically, We draw the neighboring connected landmark points, with each connection encoded in different colors as shown in Figure~\ref{fig:overall_framework}(a). 

One can thus take the landmark images of the source and driving, along with the source image, \ie, $\bm{x}_{\textnormal{in}} = \textnormal{Concat}({\bm{I}}_s, {\bm{I}}_s^{\textnormal{ldmk}}, \bm{I}_{d}^{\textnormal{ldmk}})$, for warping field  prediction. To ensure a globally coherent prediction, we strengthen the warping capability using the condition of a latent motion code $\bm{z}_{w}$ which is derived from the input, \ie,
$\bm{z}_{w} = E_{w}(\bm{x}_{\textnormal{in}})$, where $E_{w}$ is a CNN encoder. The motion code $\bm{z}_{w}$ is injected into the flow estimation network $F_w$ through AdaIN~\cite{XunHuang2017ArbitraryST}. By modulating the mean and variance of the normalized feature map, the network could be effectively guided by the motion vector which induces a globally coherent flow prediction. Formally, we obtain the prediction of a dense flow field through
$\bm{w} = F_w(\bm{x}_{\textnormal{in}}, \bm{z}_w).
$

\noindent\textbf{ID-preserving refinement network.} Directly warping the source image with the predicted flow field inevitably introduces artifacts and the loss of subtle perceived identity. Therefore, an ID-preserving refinement network is needed to produce a photo-realistic result while maintaining the identity of the source image. Prior works primarily focus on the geometry-aware flow field prediction, whereas such deformation may not well characterize the fine-grained facial geometry. In this work, we resolve the identity loss via a well-designed identity-preserving refinement network.

We propose to attentionally incorporate the semantic identity vector with the intermediate warping results. Let $\bm{h}_{\textnormal{in}}^{l}$ be the $l$-th layer feature map of the refinement network. We obtain the identity-aware feature output $\bm{h}_{\textnormal{id}}^{l}$ by modulating $\bm{h}_{\textnormal{in}}^{l}$ with the identity embedding $\bm{z}_{\textnormal{id}}$ through AdaIN, where $\bm{z}_{\textnormal{id}}$ is extracted using a pre-trained face recognition model $E_{\textnormal{id}}$~\cite{deng2020retinaface}. Meanwhile, we obtain a motion-aware feature  $\bm{h}_{\textnormal{motion}}^{l}$, which keeps the head motion and expression of the driving video. Specifically, $\bm{h}_{\textnormal{motion}}^{l}$ is obtained via a Feature-wise Linear Modulate (FiLM)~\cite{dumoulin2018feature-wise,perez2018film,park2019SPADE} according to the warped image $\tilde{\bm{I}}_s=\bm{w}(\bm{I}_s)$, \ie, $\bm{h}_{\textnormal{motion}}^{l} = \textnormal{Conv}(\tilde{\bm{I}}_s)\times \bm{h}_{\textnormal{id}}^l + \textnormal{Conv}(\tilde{\bm{I}}_s)$.

With both the identity-aware and motion-aware features, we adaptively fuse the features through an attention-based fusion block. Inspired by recent face-swapping approaches~\cite{LingzhiLi2019FaceShifterTH}, we suppose that the driving motions and source identity should be fused in a spatially-adaptive manner, in which the facial parts that mostly characterize the key facial features express the identity should rely more on the identity-aware feature, whereas other parts (\eg hair and clothes) should make greater use of the motion-aware feature. A learnable fusion mask $\bm{M}^l$ is used for the fusion of these two parts, which is predicted by,
\begin{equation}
    \bm{M}^l = \mathbf{\sigma}(\textnormal{Conv}(\bm{h}_{\textnormal{in}}^l)),
\end{equation}
where $\mathbf{\sigma}$ indicates the sigmoid activation function. In this way, the model learns to properly inject the identity-aware features into identity-related regions. 
The output of layer $l$ can be derived by fusing features according to the mask $\bm{M}^l$, which is,
\begin{equation}
    \bm{h}_{\textnormal{out}}^l = \bm{M}^l\otimes\bm{h}_{\textnormal{motion}}^{l} + (1 - \bm{M}^l)\otimes\bm{h}_{\textnormal{id}}^{l},
\end{equation}
where $\otimes$ denotes the Hadamard product. Through a cascade of such blocks, we obtain the final output image $\tilde{{\bm{I}}}_y$, which well preserves the source identity while accurately following the head motion and expression as the driving person.

\noindent\textbf{Training objective.} Perceptual loss~\cite{johnson2016perceptual} is computed between the warped source image $\tilde{{\bm{I}}}_s$ and ground-truth driving image ${\bm{I}}_d$ for accurate warping prediction. The same loss is also applied to enforce the refinement output $\tilde{{\bm{I}}}_y$.

We also extract the feature using face recognition model $E_{\textnormal{id}}$~\cite{deng2020retinaface}, and penalize the dissimilarity between the ID vectors of the output image $\tilde{{{\bm{I}}}}_y$ and source ${\bm{I}}_s$, using
\begin{equation}
\mathcal{L}_{\textnormal{id}} = 1 - \cos{(E_{\textnormal{id}}(\tilde{{\bm{I}}}_y), E_{\textnormal{id}}({\bm{I}}_s))}.
\end{equation}
A multi-scale patch discriminator $\mathcal{L}_{\textnormal{adv}}$~\cite{park2019SPADE} is adopted to enhance the photo-realism of the outputs. To further improve the generation quality on the hard eye and mouth areas, we add additional $\mathcal{L}_1$ reconstruction losses, \ie, $\mathcal{L}_{\textnormal{eye}}, \mathcal{L}_{\textnormal{mouth}}$, for these parts. 

The overall training loss can be formulated as
\begin{equation}
\begin{split}
    \mathcal{L} = & \mathcal{L}_{\textnormal{w}}^{\textnormal{VGG}} + \lambda_{\textnormal{r}}\mathcal{L}_{\textnormal{r}}^{\textnormal{VGG}} + \lambda_{\textnormal{id}}\mathcal{L}_{\textnormal{id}} + \\ &\lambda_{\textnormal{eye}}\mathcal{L}_{\textnormal{eye}} + \lambda_{\textnormal{mouth}}\mathcal{L}_{\textnormal{mouth}} +
    \lambda_{\textnormal{adv}}\mathcal{L}_{\textnormal{adv}},
\end{split}
\end{equation}
where $\lambda_{\textnormal{r}}, \lambda_{\textnormal{id}}, \lambda_{\textnormal{eye}}, \lambda_{\textnormal{mouth}}$, and $\lambda_{\textnormal{adv}}$ are the loss weights.

\subsection{Meta-learning based Faster Personalization}
\label{sec:meta}

\begin{figure}[t]
    \centering
    \begin{overpic}
    [width=0.85\columnwidth]{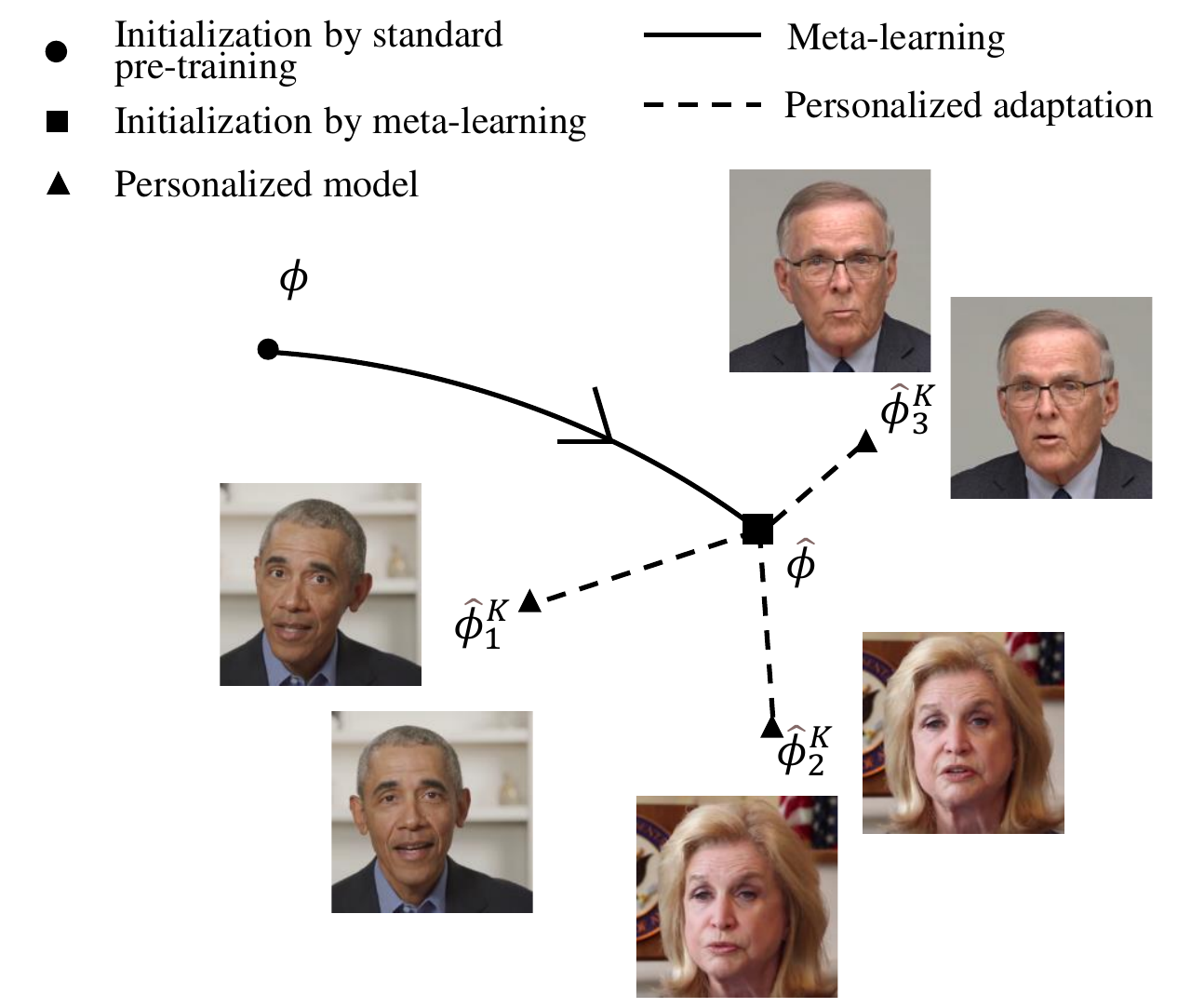}
    \end{overpic}
    \caption{Visualization of meta-learning. Compared with the one-shot pre-trained model $\bm{\phi}$,  the meta-learned model $\hat{\bm{\phi}}$ can be rapidly adapted to a unique personal model $\hat{\bm{\phi}}_j^K$ in $K$ steps.
    }    
    \label{fig:MAML}
    \vspace{-0.5cm}
\end{figure}

While achieving state-of-the-art generation quality using our one-shot model, there always exists challenging cases that are intractable since the one-shot setting is inherently ill-posed. 
Person-specific characteristics and occlusion would never be faithfully recovered using a one-shot general model. Thus, personalized fine-tuning is necessary to
achieve robust and authentic results that are truly usable. 
Nonetheless, fine-tuning the one-shot pre-trained model on a long video is computationally prohibitive for common users.
To solve this, we propose a meta-learned model whose initialization weights can be easily adapted according to low-shot personal data within a few training steps, as illustrated in Figure~\ref{fig:MAML}.

Formally, given a person $j$, the personalized adaptation starts from the pre-trained model weight $\bm{\phi}$ and aims to reach the optimal personal model weight $\hat{\bm{\phi}}_{j}$ by minimizing the error against the personal images $\hat{\bm{X}}_j$:
\begin{equation}
    \hat{\bm{\phi}}_{j} = \min_{\bm{\phi}_j}\mathcal{L}(G_{\bm{\phi}_j}(\hat{\bm{X}}_j)),
\end{equation}
where $G_{{\bm{\phi}_j}}$ denotes the whole generator with weight $\phi_j$.
Usually we perform $K$ steps of stochastic gradient descent (SGD) from initialization $\bm{\phi}$ to approach $\hat{\bm{\phi}}_{j}$, so the weight updating process can be formulated as:
\begin{equation}
    \bm{\phi}_j^k = \textnormal{SGD}_K(\bm{\phi}, \hat{\bm{X}}_j).
\end{equation}
Our goal is to find an optimal initialization $\hat{\bm{\phi}}^K$ which could approach any personal model after $K$ steps of SGD update, even for a small $K$, \ie, 
\begin{equation}
    \hat{\bm{\phi}}^K = \min_{\bm{\phi}}\sum_{j=1}^{M}\|\hat{\bm{\phi}}_j - \bm{\phi}_j^k\|.
\label{eq:min}
\end{equation}

Indeed, the general one-shot pretrained model $\bm{\phi}$ is a special case in the above formulation, which essentially learns the model weight in Equation~\ref{eq:min} when $K=0$. When we are allowed to perform a few adaption steps ($K>0$), there is a gap between $\bm{\phi}$ and desired $\hat{\bm{\phi}}^{K}$, since the optimization target of the standard pre-training is to minimize the overall error across all training data, it does not necessarily find the best weight suitable for personalization.

Compared with general one-shot models, we leverage the idea of Model-Agnostic Meta-Learning (MAML) to bridge this gap and enable surprisingly fast personalized training. The goal of MAML-based methods is to optimize the initialized weights such that they could be fast adapted to a new identity within a few steps of gradient descent, which directly matches our goal. Directly optimizing the initialization weight using Equation~\eqref{eq:min} involves the computation of second-order derivatives, which is computationally expensive on large-scale training. Therefore, we utilize Reptile~\cite{AlexNichol2018ReptileAS}, a first-order MAML-based approach to obtain suitable initialization for fast personalization. 

To be more specific, our meta-learning model explicitly considers the personalized adaptation during training. For each person $j$, we start from the $\bm{\phi}_j^0=\bm{\phi}$, which is the initialization to be optimized. Formally, we sample a batch of personal training data $\hat{\bm{X}}_j$, the $K$ steps of personalized training yield the finetuned model weights as:
\begin{equation}
\bm{\phi}_j^k=\textnormal{SGD}\left(\bm{\phi}_j^{k-1},\hat{\boldsymbol{X}}_j\right), \textnormal{ }{k = 1, \cdots, K}.
\end{equation}
Finally, the personal update, \ie, the difference of $\bm{\phi}_j^K$ and $\bm{\phi}_j^0$, is used as the gradient to update our initialization $\bm{\phi}$:
\begin{equation}
    \bm{\phi} \leftarrow \bm{\phi}-{\beta}\left(\bm{\phi}_j^K-\bm{\phi}\right),
\end{equation}
where $\beta$ is the meta-learning rate. The full algorithm is shown in Algorithm~\ref{alg:reptile}. Our model progressively learns a more suitable initialization through meta-training as visualized in Figure~\ref{fig:MAML}, which could fast adapt to personalized models after limited steps of adaptation.

\begin{algorithm}[t]
\small
\caption{Optimization of Initial Weights with Reptile}\label{alg:reptile}
\begin{algorithmic}[1]
\Statex \textbf{Input}: weights $\bm{\phi}$ of generation network $G$, inner loop learning rate $\alpha$, meta-learning rate $\beta$, number of training iterations $N$, 
number of training persons $M$,
number of inner loop iterations $K$
\For{$i = 1, \cdots, N$}
    \For{$j = 1, \cdots, M$}
        \State $\bm{\phi}_j^{0} = \textbf{Clone}(\bm{\phi})$
        \State Sample a training batch $\hat{\bm{X}}_j$ of person $j$ 
        \For{$k = 1, \cdots, K$}
            \State $\bm{\phi}_j^{k}=\bm{\phi}_{j}^{k-1}-\alpha \nabla_{\bm{\phi}} \mathcal{L}(G_{\bm{\phi}}(\hat{\bm{X}}_j))$
        \EndFor
    \EndFor
    \State $\bm{\phi} \leftarrow \bm{\phi}-\frac{\beta}{M}\sum_{j=1}^M (\bm{\phi}_{j}^{K} - \bm{\phi}$)
\EndFor
\end{algorithmic}
\end{algorithm}

\begin{figure*}[t]
    \center
    \footnotesize
    \setlength\tabcolsep{1pt}
    {
    \renewcommand{\arraystretch}{0.6}
    \begin{tabular}{@{}ccccccc@{}}
         \includegraphics[width=0.28\columnwidth]{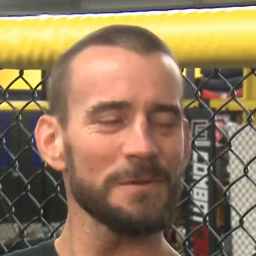} & \includegraphics[width=0.28\columnwidth]{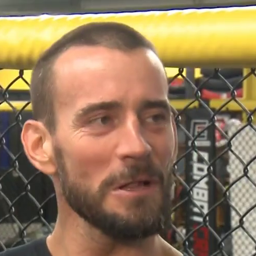} & \includegraphics[width=0.28\columnwidth]{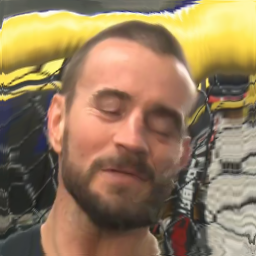} & \includegraphics[width=0.28\columnwidth]{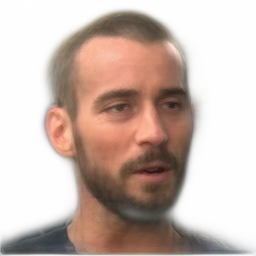} & \includegraphics[width=0.28\columnwidth]{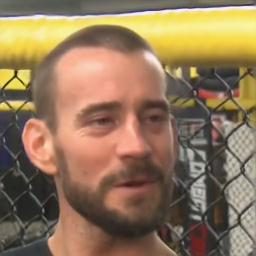} & \includegraphics[width=0.28\columnwidth]{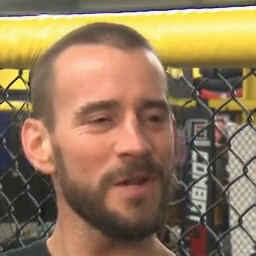} & \includegraphics[width=0.28\columnwidth]{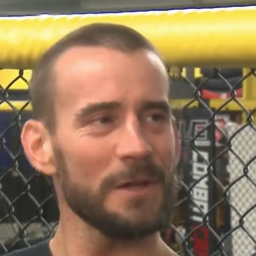}\\
         \includegraphics[width=0.28\columnwidth]{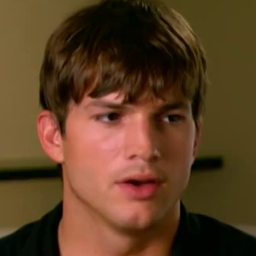} & \includegraphics[width=0.28\columnwidth]{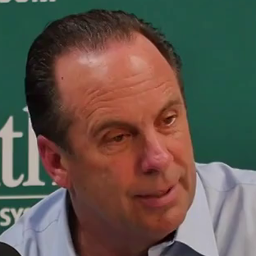} & \includegraphics[width=0.28\columnwidth]{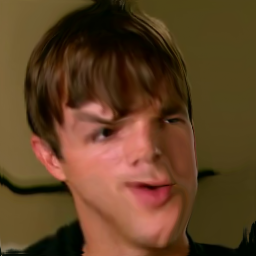} & \includegraphics[width=0.28\columnwidth]{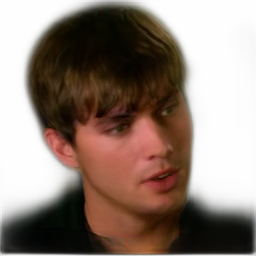} & \includegraphics[width=0.28\columnwidth]{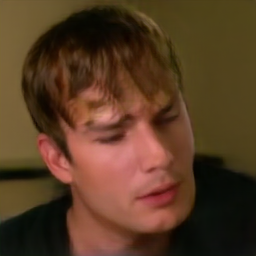} & \includegraphics[width=0.28\columnwidth]{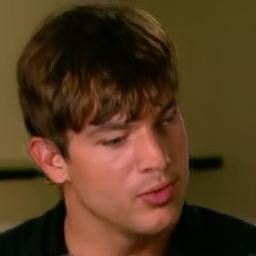} & \includegraphics[width=0.28\columnwidth]{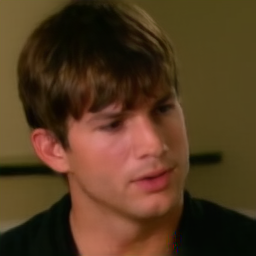}\\
         \includegraphics[width=0.28\columnwidth]{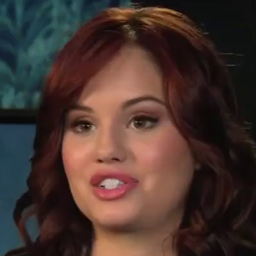} & \includegraphics[width=0.28\columnwidth]{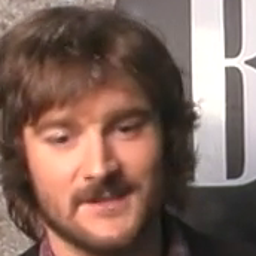} & \includegraphics[width=0.28\columnwidth]{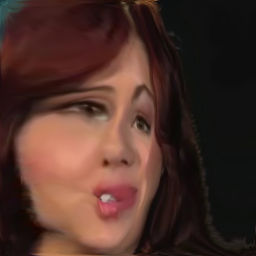} & \includegraphics[width=0.28\columnwidth]{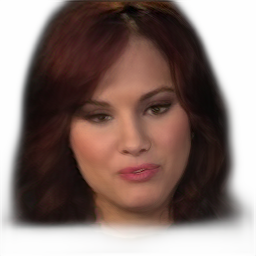} & \includegraphics[width=0.28\columnwidth]{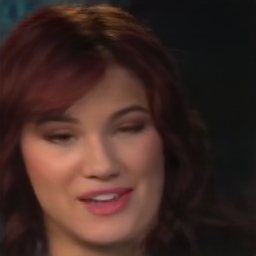} & \includegraphics[width=0.28\columnwidth]{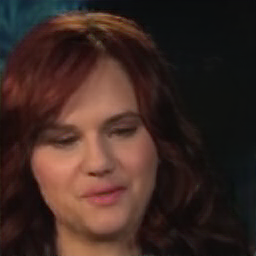} & \includegraphics[width=0.28\columnwidth]{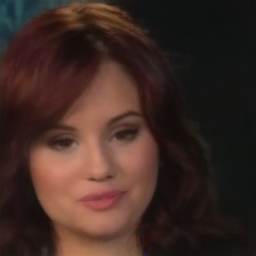}\\
          & & & & & &\\
         Source & Driving & X2Face & Bi-layer & FOMM & PIRender & Ours
    \end{tabular}
    }
    \caption{Qualitative results of our self-reconstruction (top row) and cross-identity reenactment (bottom two rows) at $256\times256$ resolution. Our method synthesizes more faithful expression and motion, while better preserving the identity of the source portrait. }
    \label{fig:general_results}
\end{figure*}

\subsection{Temporal-consistent Super-resolution Network}
\label{sec:sr}
To further enhance the generation resolution and improve the high-fidelity details of our output, a video super-resolution network is needed as the final stage of the generation framework. Previous talking head synthesis works~\cite{FeiYin2022StyleHEATOH,Drobyshev22MegaPortraits} utilize single-frame super-resolution as the last stage and ignore the quality of temporal consistency and stability. Performing super-resolution in frame-by-frame manner tends to produce texture flickering which severely hampers the visual quality. In this work, we consider multiple adjacent frames to ensure temporal coherency. 

Inspired by previous 2D face restoration works~\cite{wang2021gfpgan,TaoYang2021GANPE}, the pre-trained generative models~\cite{karras2019stylebased,Karras2019stylegan2,zhang2021styleswin} like StyleGAN contain rich face prior and could significantly help to enhance the high-frequency details. Moreover, the disentangled $\mathcal{W}$ space in StyleGAN provides desirable temporal consistency during manipulation~\cite{tzaban2022stitch}, which also benefits our framework. Therefore, we propose a temporally consistent super-resolution module by leveraging pretrained StyleGAN and 3D convolution, where the latter brings quality enhancement in spatio-temporal domain.
As shown in Fig.~\ref{fig:overall_framework}, we feed the concatenated sequence of $t$ video frames $\{\tilde{{\bm{I}}}_y^{1}, \tilde{{\bm{I}}}_y^{2}, ..., \tilde{{\bm{I}}}_y^{t}\}$ into a U-Net composed of 3D convolution with reflection padding on the temporal dimension.
To ensure pre-trained per-frame quality while improving temporal consistency, we initialize the 3D convolution weight in U-Net with a pretrained 2D face restoration network~\cite{wang2021gfpgan}.

These spatio-temporally enhanced features from the U-Net decoder further modulate the pretrained StyleGAN features through FiLM. Thus, the super-resolution frames $\{{{\bm{I}}'}_y^{1}, {{\bm{I}}'}_y^{2}, ..., {{\bm{I}}'}_y^{t}\}$ are obtained as:
\begin{equation}
    \{{{\bm{I}}'}_y^{1}, {{\bm{I}}'}_y^{2}, ..., {{\bm{I}}'}_y^{t}\} = F_{\textnormal{StyleGAN}}(F_{\textnormal{3D}}(\{\tilde{{\bm{I}}}_y^{1}, \tilde{{\bm{I}}}_y^{2}, ..., \tilde{{\bm{I}}}_y^{t}\})).
\end{equation}

During training, we optimize the output ${{\bm{I}}'}_{y}$ towards the $512\times512$ ground truth ${{\bm{I}}}_{d}$ using $\ell_1$ and perceptual loss.

\begin{figure*}[t]
    \centering
    \footnotesize
    \setlength\tabcolsep{1pt}
    {
    \renewcommand{\arraystretch}{0.6}
    \begin{tabular}{@{}ccccc@{}}
         \includegraphics[width=0.4\columnwidth]{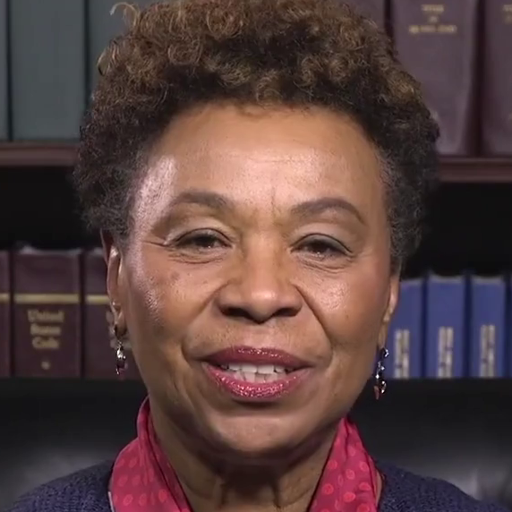} & \includegraphics[width=0.4\columnwidth]{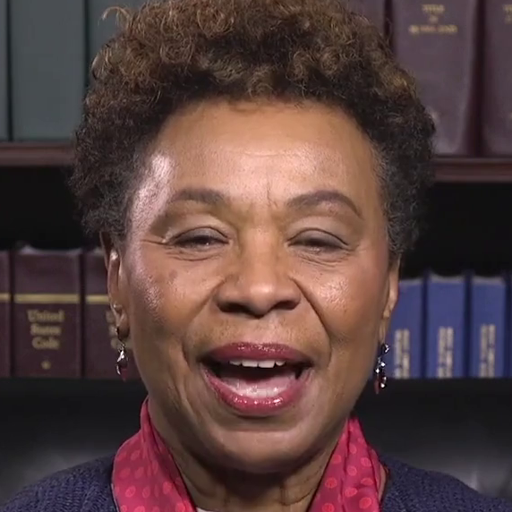} & \includegraphics[width=0.4\columnwidth]{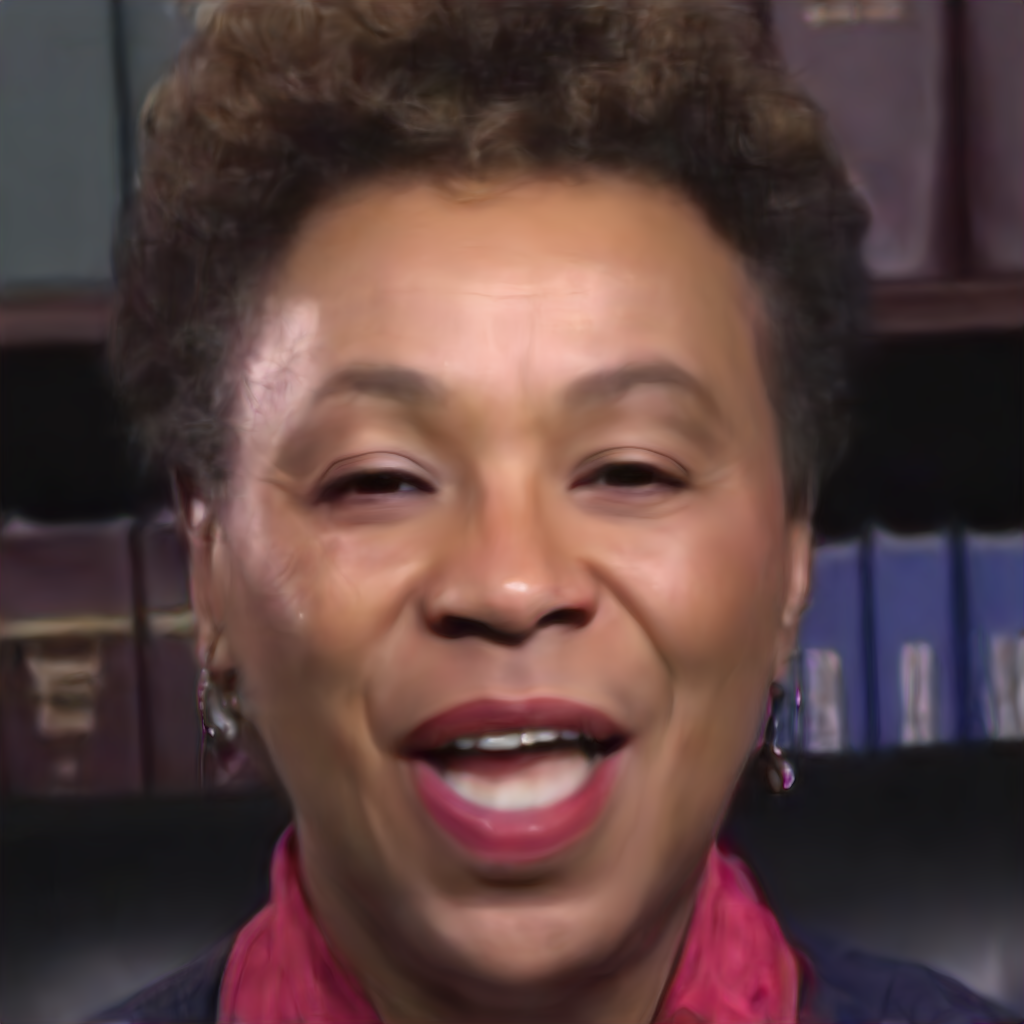} & \includegraphics[width=0.4\columnwidth]{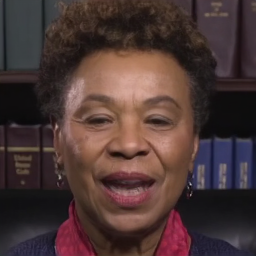} & \includegraphics[width=0.4\columnwidth]{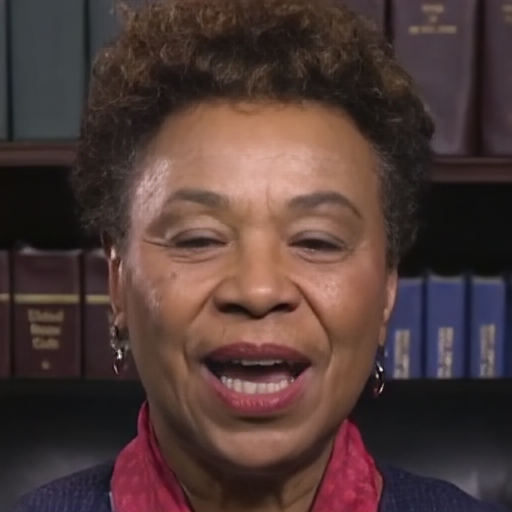}\\
         \includegraphics[width=0.4\columnwidth]{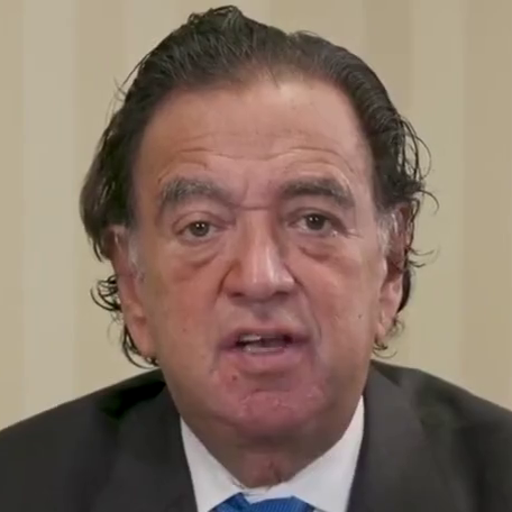} & \includegraphics[width=0.4\columnwidth]{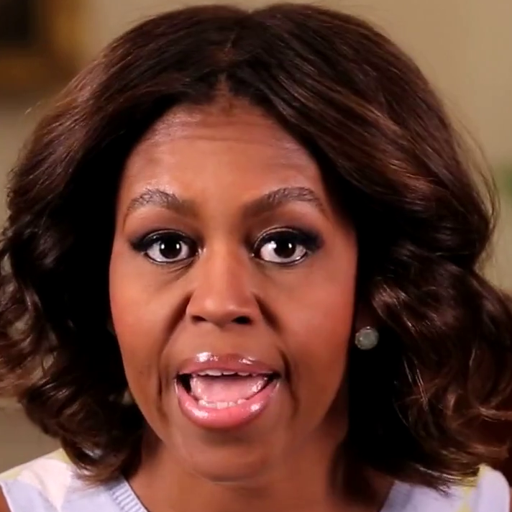} & \includegraphics[width=0.4\columnwidth]{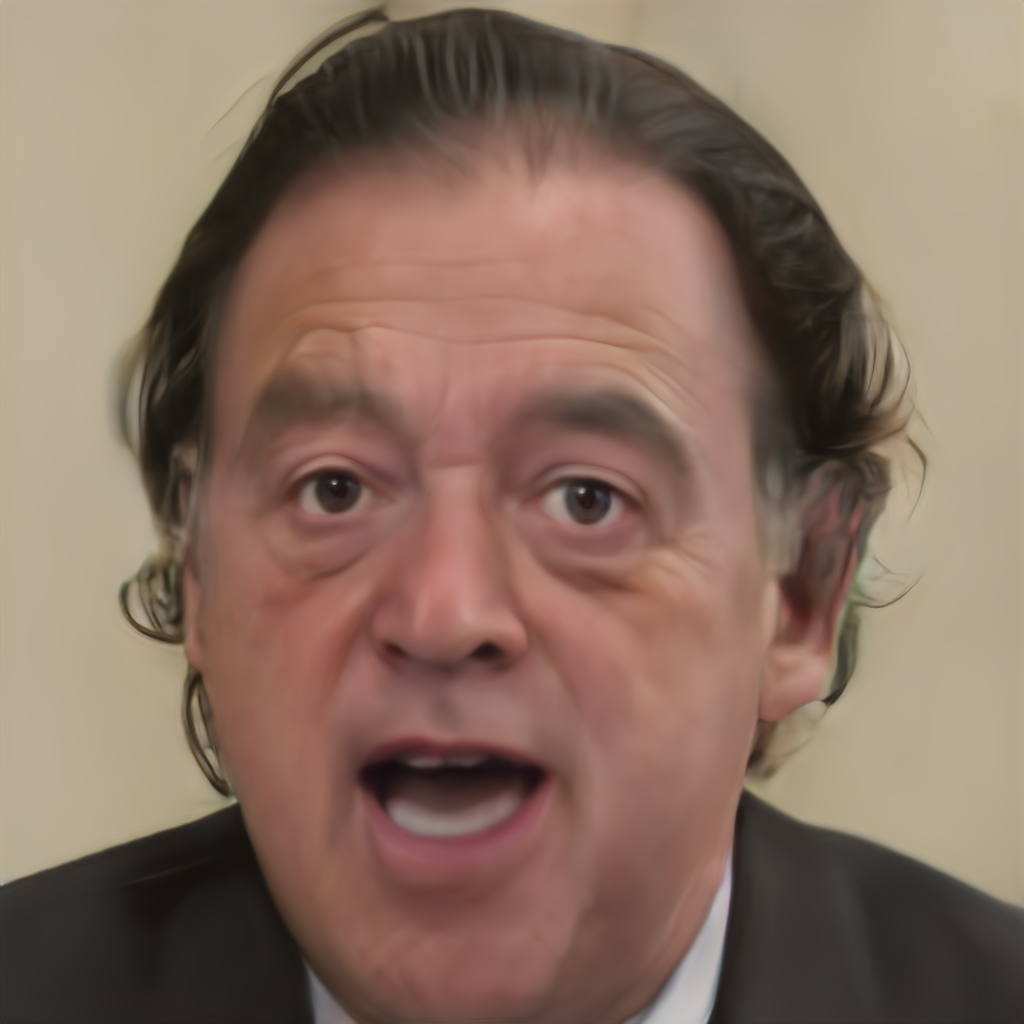} & \includegraphics[width=0.4\columnwidth]{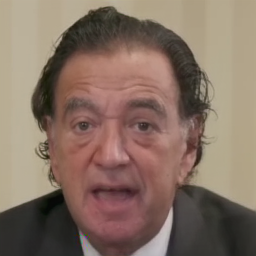} & \includegraphics[width=0.4\columnwidth]{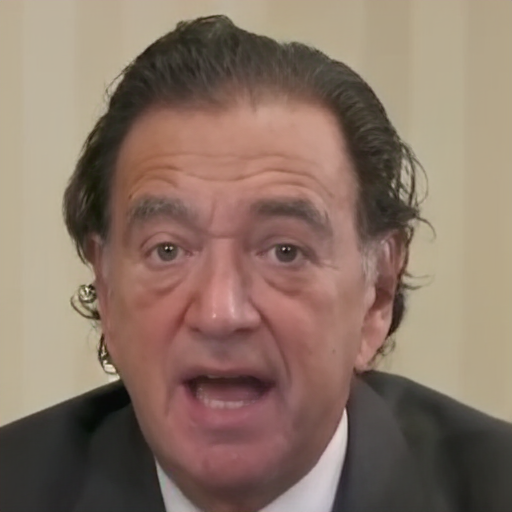}\\
          & & & & \\
         Source & Driving & StyleHEAT &  Base w/ bicubic
         & Ours
    \end{tabular}
    }
    \caption{Qualitative results of self reconstruction and cross-identity reenactment at $512\times512$ resolution. Our spatial-temporal super-resolution module further enhances more high-frequency details on teeth, eyes, and hair.}
    \label{fig:512_results}
\end{figure*}

\begin{table*}[t]
    \footnotesize
    \centering
    \begin{tabular}{c|c|c|c|c|c|c|c|c|c}
        \toprule
        \multirow{2}*{Methods} & \multicolumn{5}{c|}{\textbf{Self Reconstruction ($256 \times 256$)}} & \multicolumn{4}{c}{\textbf{Cross Reenactment ($256 \times 256$)}} \\
        \cmidrule{2-10}
        ~ & FID$\downarrow$ & LPIPS$\downarrow$ & ID Loss$\downarrow$ & AED$\downarrow$ & APD$\downarrow$ & FID$\downarrow$ & ID Loss$\downarrow$ & AED$\downarrow$ & APD$\downarrow$ \\
        \midrule
        X2Face\cite{wiles2018x2face} & 45.2908 & 0.6806 & 0.9632 & 0.2147 & 0.1007 & 91.1485 & 0.6496 & 0.3112 & 0.1210 \\
        Bi-layer\cite{EgorZakharov2020FastBN} & 100.9196 & 0.5881 & 0.5280 & 0.1258 & 0.0139 & 127.7823 & 0.6336 & 0.2330 & \textbf{0.0208} \\
        FOMM\cite{AliaksandrSiarohin2019FirstOM} & 12.1979 & 0.2338 & 0.2096 & 0.0964 & \textbf{0.0100} & 80.1637 & 0.5760 & 0.2340 & 0.0239\\
        PIRender\cite{YuruiRen2021PIRendererCP} & 14.4065 & 0.2639 & 0.3024  & 0.1080 & 0.0162 & 78.8430 & 0.5440 & \textbf{0.2113} & 0.0214\\
        \cellcolor{gray}\emph{Ours} &\cellcolor{gray}\textbf{11.9528}  & \cellcolor{gray}\textbf{0.2262} & \cellcolor{gray}\textbf{0.1296} & \cellcolor{gray}\textbf{0.0942} & \cellcolor{gray}0.0124 & \cellcolor{gray}\textbf{77.5048} & \cellcolor{gray}\textbf{0.2944} & \cellcolor{gray}0.2524 & \cellcolor{gray}0.0258 \\
        \midrule
        {Methods} & \multicolumn{5}{c}{\textbf{Self Reconstruction ($512 \times 512$)}} & \multicolumn{4}{|c}{\textbf{Cross Reenactment ($512 \times 512$)}} \\
        \midrule
        StyleHEAT\cite{FeiYin2022StyleHEATOH} & 44.5207 & 0.2840 & 0.4112  & 0.1155 & 0.0131 & 111.3450 & 0.4720 & \textbf{0.2505} & \textbf{0.0218}\\
        \cellcolor{gray}\emph{Ours}
        &\cellcolor{gray}\textbf{21.4974}  & \cellcolor{gray}\textbf{0.2079} & \cellcolor{gray}\textbf{0.0832} & \cellcolor{gray}\textbf{0.0904} & \cellcolor{gray}\textbf{0.0121} & \cellcolor{gray}\textbf{49.6020} & \cellcolor{gray}\textbf{0.1952} & 
        \cellcolor{gray}0.2737 & 
        \cellcolor{gray}0.0242 \\
        \bottomrule
    \end{tabular}
    \caption{Quantitative results for self-reconstruction and cross-reenactment. We evaluate both $256\times256$ results and $512\times512$ results. Our method outperforms all baselines on both resolutions across image fidelity metrics with comparable motion transfer results.}
    \label{tab:Quantitative_main_result}
\end{table*}

\section{Experiments}
\label{sec:exp}

\begin{figure*}[t]
    \centering
    \footnotesize
    \setlength\tabcolsep{1pt}
    \renewcommand{\arraystretch}{0.6}
     \includegraphics[width=2.1\columnwidth]{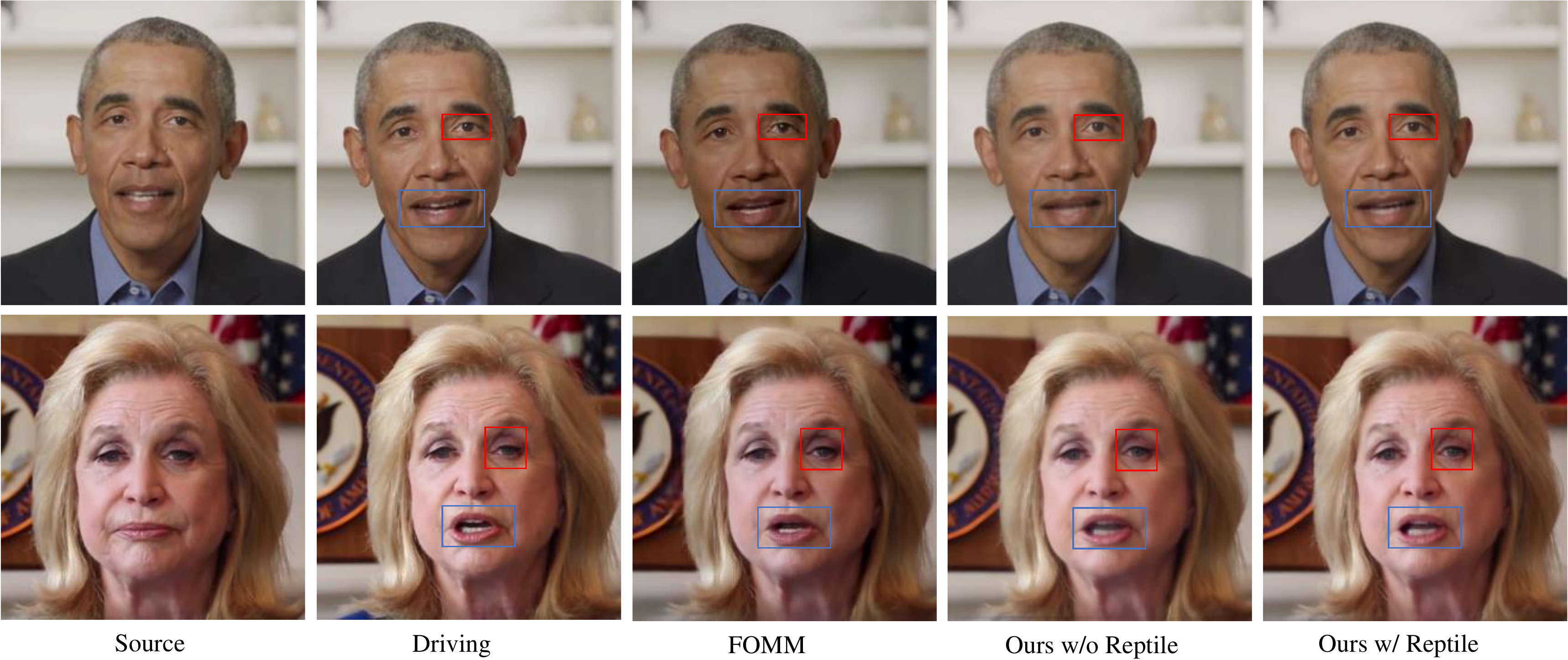} 
    \caption{A comparison of different personalized models at the same epochs. Our personalized base model without reptile adaptation reconstructs more accurate skin color and source identity than the personalized FOMM method. For the model personalized from a reptile learning stage, finer detail on teeth and eye colors could be further generated.}
    \label{fig:personalization}
\end{figure*}

\subsection{Experiment Setup}
\noindent\textbf{Dataset.}
Following~\cite{Drobyshev22MegaPortraits}, we train our warping and refinement networks on cropped VoxCeleb2 dataset~\cite{JoonSonChung2018VoxCeleb2DS} at $256^2$ resolution.
We randomly select 500 videos from the test set for evaluation. 
For our meta-learning-based fast personalization, we finetune our 
base model
on HDTF dataset~\cite{ZhimengZhang2022FlowguidedOT}, which is composed of 410 videos from 300 different identities. We downsample cropped faces to $256^2$ resolution and split the original HDTF dataset into 400 training videos and 10 test videos.
After the convergence of our meta-training, we further evaluate the personalization speed of our model on the HDTF test set. 
Our temporal super-resolution module $F_{3d}$ is trained on the HDTF dataset~\cite{ZhimengZhang2022FlowguidedOT} which has 300 frames per video with $512^2$ resolution.
We feed downsampled $256^2$ frames into fixed warping and refinement network and use the outputs of the refinement network as inputs for the next temporal super-resolution module. More training details are provided in the supplementary.

\noindent\textbf{Metrics}
We evaluate the fidelity of self-reconstruction using  FID~\cite{Heusel2017TTURFID} and LPIPS~\cite{RichardZhang2018LPIPS}. Our motion transfer quality is measured using average expression distance (AED) and average pose distance (APD) with driving videos. 
For our meta-learning-based fast personalization, we illustrate the LPIPS at different personalization epochs for each fine-tuning approach.
Following previous works~\cite{ChenyangLei2020BlindVT,Lai2018LearningBVTC}, we evaluate our temporal consistency using warping error $E_{\text {warp }}$. For each frame $\tilde{{\bm{I}}}_y^{t}$, we calculate the warping error with the previous frame $\tilde{{\bm{I}}}_y^{t-1}$ warped by the estimated optical flow in the occlusion map.

\subsection{Comparison with State-of-the-art Methods}
We compare our warping and refinement models at $256\times256$ resolution against several state-of-the-art face reenactment works: X2Face~\cite{wiles2018x2face}, Bi-Layer~\cite{ba2016layer}, First-Order Motion Model~\cite{AliaksandrSiarohin2019FirstOM}, and PIRender~\cite{YuruiRen2021PIRendererCP}. 
The top row of Figure~\ref{fig:general_results} presents our qualitative results of self-reconstruction. 
Different from sparse landmarks in unsupervised learning~\cite{AliaksandrSiarohin2019FirstOM} or 1D 3DMM~\cite{YuruiRen2021PIRendererCP}, our dense landmarks provide strong guidance to accurate and fine-detailed synthesis on gaze, mouth, and expressions. The last two rows show the results of our cross-identity reenactment. Since our landmarks have a better decomposition of identity and motion and our refinement network is identity-aware, our method is the only one that well preserves the identity of source image. In contrast, previous methods suffer from appearance data leakage directly from the driver and generate faces with a similar identity to the driving image. 
Note that the 3DMM coefficients of AED and APD evaluation are not fully decomposed. Thus, other baselines with identity leakage may have better quantitative metrics.
We compare our full framework with the proposed temporal super-resolution module at $512\times512$ resolution against StyleHEAT~\cite{FeiYin2022StyleHEATOH}, which is the only open-sourced method that generates high-resolution talking heads. 
StyleHEAT~\cite{FeiYin2022StyleHEATOH} fails to synthesize sharp and accurate teeth in our experiment, and the identity of the output image is quite different from the source portrait due to the limitations of GAN inversion. In contrast, our refinement network is identity-aware and we leverage pretrained StyleGAN in our temporal super-resolution module to fully exploit its face prior knowledge. Figure~\ref{fig:512_results} shows that our method produces sharp teeth and hairs, while bicubic-upsampled results are blurry with artifacts.

Table~\ref{tab:Quantitative_main_result} demonstrates that our method achieves the best quantitative fidelity and comparable motion transfer quality on both self-reconstruction and cross-identity reenactment.

\definecolor{myblue}{RGB}{0, 28, 127}
\definecolor{mygreen}{RGB}{18, 113, 28}
\definecolor{myorange}{RGB}{177, 63, 13}
\pgfplotsset{grid style={dashed}}
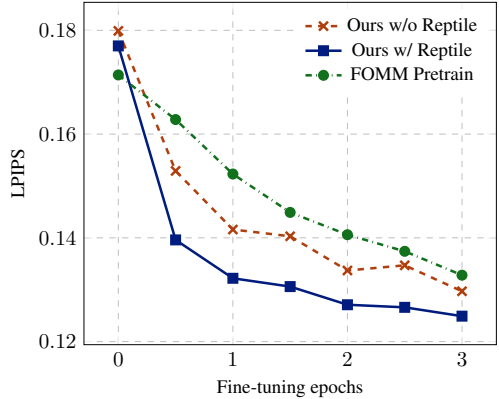
\begin{figure}[t]
    \centering
    \begin{tikzpicture}[scale=0.8]
    \begin{axis}[
    	xlabel = {Fine-tuning epochs},
    	ylabel = {LPIPS},
    	label style={font=\small},
    	grid = major,
    	/pgfplots/xtick = {0,1,2,3},
    	legend entries = {Ours w/o Reptile, Ours w/ Reptile, FOMM Pretrain},
    	legend style = {draw=none, font=\small},
    	]
    	\addplot [mark=x, mark options={scale=1.5, solid}, dashed, myorange, very thick] table {images/results/meta/baseline.dat};
    	\addplot [mark=square*, very thick, myblue]table {images/results/meta/reptile.dat};
    	\addplot [mark=otimes*, mark options={scale=1, solid}, dashdotted, very thick, mygreen]table {images/results/meta/fomm.dat};
    \end{axis}
\end{tikzpicture}

\caption{Comparison of test LPIPS at fine-tuning epochs for different approaches. Our reptile-based model achieves more than $3\times$ speedup compared with the base model and previous baseline.}
\label{fig:adaptiona_curve}
\end{figure}

\subsection{Evaluation of Fast Personalization}

Although our base model achieves state-of-the-art quality as a general model, there still exists ill-posedness in some cases. For example, it is very difficult to synthesize teeth according to a closed source mouth. Thus, industry products typically conduct personalization using fine-tuning strategy, which can be computationally expensive. To achieve faster convergence, we finetune our base model using a meta-learning strategy to provide a better weight initialization for the following personalization. In Fig~\ref{fig:adaptiona_curve}, we evaluate the personalization speed of our meta-learned model against our base model and previous baseline FOMM~\cite{AliaksandrSiarohin2019FirstOM}. It takes our meta-learned model 0.5 epoch to decrease LPIPS to 0.14, which is $3\times$ speedup against our base model, and $4\times$ against FOMM. Figure~\ref{fig:personalization} compares the personalization of our method and FOMM~\cite{AliaksandrSiarohin2019FirstOM} at the same epoch, which illustrates our fast adaptation speed on ambiguous areas (\eg, teeth, eyes, and wrinkle details).

\begin{table}[t]
    \footnotesize
    \centering
    \begin{tabular}{l|c|c|c}
        \toprule
        {Methods} & FID$\downarrow$ & LPIPS$\downarrow$  & $E_{warp}\downarrow$ \\
        \midrule
        Ground Truth & - & -   & \textbf{0.0182}  \\
        \midrule
Base w/ bicubic & 25.5762 & 0.2285   & 0.0184  \\
        Base w/ GFPGAN~\cite{wang2021gfpgan} & 22.6351 & 0.2178   & 0.0242  \\
        \cellcolor{gray}\emph{Ours}
        &\cellcolor{gray}\textbf{21.4974}  & \cellcolor{gray}\textbf{0.2079} & \cellcolor{gray}0.0213   \\
        \bottomrule
    \end{tabular}
    \caption{Quantitative evaluation of our temporal super-resolution on self-reconstruction at $512\times512$ resolution.}
    \label{table: temporal_consistency}
\end{table}

\subsection{Evaluation of Temporal Super-resolution}
In Table~\ref{table: temporal_consistency}, we also evaluate the performance of our temporal super-resolution using 2D image fidelity and warping error $E_{warp}$. 
We train a 2D super-resolution baseline using GFPGAN~\cite{wang2021gfpgan}.
The quantitative result in Table~\ref{table: temporal_consistency} shows that although naive 2D super-resolution improves per-frame fidelity, it also brings more flickering and larger warping error (0.0242) than simple bicubic upsampling (0.0184). To achieve temporally coherent results, we combine a U-Net composed of 3D convolution with facial prior~\cite{Karras2019stylegan2}, which significantly reduces $E_{warp}$ of our final videos from 0.0242 to 0.0213, and preserves compelling 2D facial details.

\subsection{Ablation Study of Base Model}

\begin{table}[th]
    \footnotesize
    \centering
    \begin{tabular}{l|c|c|c}
        \toprule
        {Methods} & FID$\downarrow$ & LPIPS$\downarrow$ & ID Loss$\downarrow$  \\
        \midrule
        Ours Sparse Landmark & 14.3190 & 0.2485 & 0.1424  \\
        Ours w/o ID & 12.2736 & \textbf{0.2256} & 0.2144  \\
        \cellcolor{gray}\emph{Ours}
        & \cellcolor{gray} \textbf{11.9528} & \cellcolor{gray}0.2262 & \cellcolor{gray}\textbf{0.1296}   \\
        \bottomrule
    \end{tabular}
    \caption{Quantitative ablation study of landmarks and ID coefficients in our base model.}    
\end{table}

We conduct ablation studies to validate the effectiveness of our driving motion and source identity representation in our base model.
If we replace our 669 dense landmarks with sparse landmarks, the LPIPS of warped source images degrades by 0.2. To evaluate our identity-aware refinement, the removal of the identity input causes significant increase of identity loss from 0.1296 to 0.2144.

\section{Conclusion}
We present a novel framework for identity-preserving one-shot talking head generation.
To faithfully maintain the source ID, we propose to leverage accurate dense landmarks in the warping network and explicit source identity during refinement. Further, we significantly advance the applicability of personalized model by reducing its training to 30 seconds with meta-learning. Last but not least, we enhance the final resolution and temporal consistency with 3D convolution and generative prior. Comprehensive experiments demonstrate the state-of-the-art performance of our system.

{\footnotesize
\bibliographystyle{ieee_fullname}
\bibliography{egbib}

\begin{thebibliography}{10}\itemsep=-1pt

\bibitem{agarap2018deep}
Abien~Fred Agarap.
\newblock Deep learning using rectified linear units (relu).
\newblock {\em arXiv preprint arXiv:1803.08375}, 2018.

\bibitem{ba2016layer}
Jimmy~Lei Ba, Jamie~Ryan Kiros, and Geoffrey~E Hinton.
\newblock Layer normalization.
\newblock {\em arXiv preprint arXiv:1607.06450}, 2016.

\bibitem{VolkerBlanz1999A3DMM}
Volker Blanz and Thomas Vetter.
\newblock A morphable model for the synthesis of 3d faces.
\newblock {\em international conference on computer graphics and interactive
  techniques}, 1999.

\bibitem{JamesBooth2016A3DMM}
James Booth, Anastasios Roussos, Stefanos Zafeiriou, Allan Ponniahy, and David
  Dunaway.
\newblock A 3d morphable model learnt from 10,000 faces.
\newblock In {\em CVPR}, 2016.

\bibitem{JamesBooth2016A3M}
James Booth, Anastasios Roussos, Stefanos Zafeiriou, Allan Ponniahy, and David
  Dunaway.
\newblock A 3d morphable model learnt from 10,000 faces.
\newblock In {\em CVPR}, 2016.

\bibitem{JoonSonChung2018VoxCeleb2DS}
Joon~Son Chung, Arsha Nagrani, and Andrew Zisserman.
\newblock Voxceleb2: Deep speaker recognition.
\newblock {\em conference of the international speech communication
  association}, 2018.

\bibitem{deng2020retinaface}
Jiankang Deng, Jia Guo, Evangelos Ververas, Irene Kotsia, and Stefanos
  Zafeiriou.
\newblock Retinaface: Single-shot multi-level face localisation in the wild.
\newblock In {\em CVPR}, 2020.

\bibitem{doukas2020headgan}
Michail~Christos Doukas, Stefanos Zafeiriou, and Viktoriia Sharmanska.
\newblock Headgan: One-shot neural head synthesis and editing.
\newblock In {\em ICCV}, 2021.

\bibitem{Drobyshev22MegaPortraits}
Nikita Drobyshev, Jenya Chelishev, Taras Khakhulin, Aleksei Ivakhnenko, Victor
  Lempitsky, and Egor Zakharov.
\newblock Megaportraits: One-shot megapixel neural head avatars.
\newblock In {\em ACM Multimedia}, 2022.

\bibitem{dumoulin2018feature-wise}
Vincent Dumoulin, Ethan Perez, Nathan Schucher, Florian Strub, Harm~de Vries,
  Aaron Courville, and Yoshua Bengio.
\newblock Feature-wise transformations.
\newblock {\em Distill}, 2018.
\newblock https://distill.pub/2018/feature-wise-transformations.

\bibitem{ChelseaFinn2017ModelAgnosticMF}
Chelsea Finn, Pieter Abbeel, and Sergey Levine.
\newblock Model-agnostic meta-learning for fast adaptation of deep networks.
\newblock In {\em ICML}, 2017.

\bibitem{Fried2019Text-based-Editing}
Ohad Fried, Ayush Tewari, Michael Zollh\"{o}fer, Adam Finkelstein, Eli
  Shechtman, Dan~B Goldman, Kyle Genova, Zeyu Jin, Christian Theobalt, and
  Maneesh Agrawala.
\newblock Text-based editing of talking-head video.
\newblock {\em ACM Trans. Graph.}, 2019.

\bibitem{Geng2018WarpguidedGF}
Jiahao Geng, Tianjia Shao, Youyi Zheng, Yanlin Weng, and Kun Zhou.
\newblock Warp-guided gans for single-photo facial animation.
\newblock {\em ACM Transactions on Graphics (TOG)}, 2018.

\bibitem{XiaojieGuo2019PFLDAP}
Xiaojie Guo, Siyuan Li, Jiawan Zhang, Jiayi Ma, Lin Ma, Wei Liu, and Haibin
  Ling.
\newblock Pfld: A practical facial landmark detector.
\newblock {\em arXiv: Computer Vision and Pattern Recognition}, 2019.

\bibitem{ha2020marionette}
Sungjoo Ha, Martin Kersner, Beomsu Kim, Seokjun Seo, and Dongyoung Kim.
\newblock Marionette: Few-shot face reenactment preserving identity of unseen
  targets.
\newblock In {\em AAAI}, 2020.

\bibitem{Heusel2017TTURFID}
Martin Heusel, Hubert Ramsauer, Thomas Unterthiner, Bernhard Nessler, and Sepp
  Hochreiter.
\newblock Gans trained by a two time-scale update rule converge to a local nash
  equilibrium.
\newblock In {\em NeurIPS}, 2017.

\bibitem{hong2022depth}
Fa-Ting Hong, Longhao Zhang, Li Shen, and Dan Xu.
\newblock Depth-aware generative adversarial network for talking head video
  generation.
\newblock In {\em CVPR}, 2022.

\bibitem{XunHuang2017ArbitraryST}
Xun Huang and Serge Belongie.
\newblock Arbitrary style transfer in real-time with adaptive instance
  normalization.
\newblock In {\em ICCV}, 2017.

\bibitem{johnson2016perceptual}
Justin Johnson, Alexandre Alahi, and Li Fei-Fei.
\newblock Perceptual losses for real-time style transfer and super-resolution.
\newblock In {\em ECCV}, 2016.

\bibitem{karras2019stylebased}
Tero Karras, Samuli Laine, and Timo Aila.
\newblock A style-based generator architecture for generative adversarial
  networks.
\newblock In {\em CVPR}, 2019.

\bibitem{Karras2019stylegan2}
Tero Karras, Samuli Laine, Miika Aittala, Janne Hellsten, Jaakko Lehtinen, and
  Timo Aila.
\newblock Analyzing and improving the image quality of {StyleGAN}.
\newblock In {\em CVPR}, 2020.

\bibitem{Khakhulin2022ROME}
Taras Khakhulin, Vanessa Sklyarova, Victor Lempitsky, and Egor Zakharov.
\newblock Realistic one-shot mesh-based head avatars.
\newblock In {\em ECCV}, 2022.

\bibitem{kim2019neural}
Hyeongwoo Kim, Mohamed Elgharib, Hans-Peter Zoll{\"o}fer, Michael~Seidel, Thabo
  Beeler, Christian Richardt, and Christian Theobalt.
\newblock Neural style-preserving visual dubbing.
\newblock {\em ACM Transactions on Graphics (TOG)}, 2019.

\bibitem{HyeongwooKim2018DeepVP}
Hyeongwoo Kim, Pablo Garrido, Ayush Tewari, Weipeng Xu, Justus Thies, Matthias
  Nie{\ss}ner, Patrick P{\'e}rez, Christian Richardt, Michael Zollh{\"o}fer,
  and Christian Theobalt.
\newblock Deep video portraits.
\newblock {\em Association for Computing Machinery}, 2018.

\bibitem{Lai2018LearningBVTC}
Wei-Sheng Lai, Jia-Bin Huang, Oliver Wang, Eli Shechtman, Ersin Yumer, and
  Ming-Hsuan Yang.
\newblock Learning blind video temporal consistency.
\newblock In {\em ECCV}, 2018.

\bibitem{ChenyangLei2020BlindVT}
Chenyang Lei, Yazhou Xing, and Qifeng Chen.
\newblock Blind video temporal consistency via deep video prior.
\newblock In {\em NeurIPS}, 2020.

\bibitem{LingzhiLi2019FaceShifterTH}
Lingzhi Li, Jianmin Bao, Hao Yang, Dong Chen, and Fang Wen.
\newblock Faceshifter: Towards high fidelity and occlusion aware face swapping.
\newblock {\em arXiv: Computer Vision and Pattern Recognition}, 2019.

\bibitem{mallya2022implicit}
Arun Mallya, Ting-Chun Wang, and Ming-Yu Liu.
\newblock {Implicit Warping for Animation with Image Sets}.
\newblock In {\em NeurIPS}, 2022.

\bibitem{ArshaNagrani2017VoxCelebAL}
Arsha Nagrani, Joon~Son Chung, and Andrew Zisserman.
\newblock Voxceleb: A large-scale speaker identification dataset.
\newblock {\em conference of the international speech communication
  association}, 2017.

\bibitem{AlexNichol2018ReptileAS}
Alex Nichol and John Schulman.
\newblock Reptile: a scalable metalearning algorithm.
\newblock {\em arXiv: Learning}, 2018.

\bibitem{ouyang2022real}
Hao Ouyang, Bo Zhang, Pan Zhang, Hao Yang, Jiaolong Yang, Dong Chen, Qifeng
  Chen, and Fang Wen.
\newblock Real-time neural character rendering with pose-guided multiplane
  images.
\newblock {\em ECCV}, 2022.

\bibitem{pang2023dpe}
Youxin Pang, Yong Zhang, Weize Quan, Yanbo Fan, Xiaodong Cun, Ying Shan, and
  Dong-ming Yan.
\newblock Dpe: Disentanglement of pose and expression for general video
  portrait editing.
\newblock {\em arXiv preprint arXiv:2301.06281}, 2023.

\bibitem{park2019SPADE}
Taesung Park, Ming-Yu Liu, Ting-Chun Wang, and Jun-Yan Zhu.
\newblock Semantic image synthesis with spatially-adaptive normalization.
\newblock In {\em CVPR}, 2019.

\bibitem{perez2018film}
Ethan Perez, Florian Strub, Harm De~Vries, Vincent Dumoulin, and Aaron
  Courville.
\newblock Film: Visual reasoning with a general conditioning layer.
\newblock In {\em AAAI}, 2018.

\bibitem{YuruiRen2021PIRendererCP}
Yurui Ren, Ge Li, Yuanqi Chen, Thomas~H. Li, and Shan Liu.
\newblock Pirenderer: Controllable portrait image generation via semantic
  neural rendering.
\newblock In {\em ICCV}, 2021.

\bibitem{RuderDB2016VideoStyle}
Manuel Ruder, Alexey Dosovitskiy, and Thomas Brox.
\newblock Artistic style transfer for videos.
\newblock In {\em German Conference on Pattern Recognition}, pages 26--36,
  2016.

\bibitem{shin2019uncanny}
Mincheol Shin, Se~Jung Kim, and Frank Biocca.
\newblock The uncanny valley: No need for any further judgments when an avatar
  looks eerie.
\newblock {\em Computers in Human Behavior}, 94:100--109, 2019.

\bibitem{AliaksandrSiarohin2019FirstOM}
Aliaksandr Siarohin, St{\'e}phane Lathuili{\`e}re, Sergey Tulyakov, Elisa
  Ricci, and Nicu Sebe.
\newblock First order motion model for image animation.
\newblock In {\em NeurIPS}, 2019.

\bibitem{Siarohin2019MonkeyNet}
Aliaksandr Siarohin, Stéphane Lathuilière, Sergey Tulyakov, Elisa Ricci, and
  Nicu Sebe.
\newblock Animating arbitrary objects via deep motion transfer.
\newblock In {\em CVPR}, 2019.

\bibitem{tang2022explicitly}
Junshu Tang, Bo Zhang, Binxin Yang, Ting Zhang, Dong Chen, Lizhuang Ma, and
  Fang Wen.
\newblock Explicitly controllable 3d-aware portrait generation.
\newblock {\em arXiv preprint arXiv:2209.05434}, 2022.

\bibitem{tao2022structure}
Jiale Tao, Biao Wang, Borun Xu, Tiezheng Ge, Yuning Jiang, Wen Li, and Lixin
  Duan.
\newblock Structure-aware motion transfer with deformable anchor model.
\newblock In {\em CVPR}, 2022.

\bibitem{ZacharyTeed2020RAFTRA}
Zachary Teed and Jia Deng.
\newblock Raft: Recurrent all-pairs field transforms for optical flow.
\newblock In {\em ECCV}, 2020.

\bibitem{tripathy2021facegan}
Soumya Tripathy, Juho Kannala, and Esa Rahtu.
\newblock Facegan: Facial attribute controllable reenactment gan.
\newblock In {\em CVPR}, 2021.

\bibitem{tzaban2022stitch}
Rotem Tzaban, Ron Mokady, Rinon Gal, Amit~H. Bermano, and Daniel Cohen-Or.
\newblock Stitch it in time: Gan-based facial editing of real videos, 2022.

\bibitem{Unterthiner2019FVD}
Thomas Unterthiner, Sjoerd van Steenkiste, Karol Kurach, Rapha{\"{e}}l
  Marinier, Marcin Michalski, and Sylvain Gelly.
\newblock Fvd: A new metric for video generation.
\newblock In {\em ICLR}, pages 694--711. Springer, 2019.

\bibitem{Wang2021MetaAvatar}
Shaofei Wang, Marko Mihajlovic, Qianli Ma, Andreas Geiger, and Siyu Tang.
\newblock Metaavatar: Learning animatable clothed human models from few depth
  images.
\newblock In {\em NeurIPS}, 2021.

\bibitem{TingChunWang2020OneShotFN}
Ting-Chun Wang, Arun Mallya, and Ming-Yu Liu.
\newblock One-shot free-view neural talking-head synthesis for video
  conferencing.
\newblock In {\em CVPR}, 2020.

\bibitem{wang2021gfpgan}
Xintao Wang, Yu Li, Honglun Zhang, and Ying Shan.
\newblock Towards real-world blind face restoration with generative facial
  prior.
\newblock In {\em CVPR}, 2021.

\bibitem{wiles2018x2face}
Olivia Wiles, A Koepke, and Andrew Zisserman.
\newblock X2face: A network for controlling face generation using images,
  audio, and pose codes.
\newblock In {\em ECCV}, 2018.

\bibitem{ErrollWood2021FakeIT}
Erroll Wood, Tadas Baltrusaitis, Charlie Hewitt, Sebastian Dziadzio, Matthew
  Johnson, Virginia Estellers, Thomas~J. Cashman, and Jamie Shotton.
\newblock Fake it till you make it: Face analysis in the wild using synthetic
  data alone.
\newblock In {\em ICCV}, 2021.

\bibitem{wood2022dense}
Erroll Wood, Tadas Baltrusaitis, Charlie Hewitt, Matthew Johnson, Jingjing
  Shen, Nikola Milosavljevic, Daniel Wilde, Stephan Garbin, Chirag Raman, Jamie
  Shotton, Toby Sharp, Ivan Stojiljkovic, Tom Cashman, and Julien Valentin.
\newblock 3d face reconstruction with dense landmarks.
\newblock In {\em ECCV}, 2022.

\bibitem{xing2023codetalker}
Jinbo Xing, Menghan Xia, Yuechen Zhang, Xiaodong Cun, Jue Wang, and Tien-Tsin
  Wong.
\newblock Codetalker: Speech-driven 3d facial animation with discrete motion
  prior.
\newblock {\em arXiv preprint arXiv:2301.02379}, 2023.

\bibitem{https://doi.org/10.48550/arxiv.1505.00853}
Bing Xu, Naiyan Wang, Tianqi Chen, and Mu Li.
\newblock Empirical evaluation of rectified activations in convolutional
  network, 2015.

\bibitem{LingboYang2020HiFaceGANFR}
Lingbo Yang, Chang Liu, Pan Wang, Shanshe Wang, Peiran Ren, Siwei Ma, and Wen
  Gao.
\newblock Hifacegan: Face renovation via collaborative suppression and
  replenishment.
\newblock {\em ACM Multimedia}, 2020.

\bibitem{TaoYang2021GANPE}
Tao Yang, Peiran Ren, Xuansong Xie, and Lei Zhang.
\newblock Gan prior embedded network for blind face restoration in the wild.
\newblock In {\em CVPR}, 2021.

\bibitem{FeiYin2022StyleHEATOH}
Fei Yin, Yong Zhang, Xiaodong Cun, Mingdeng Cao, Yanbo Fan, Xuan Wang, Qingyan
  Bai, Baoyuan Wu, Jue Wang, and Yujiu Yang.
\newblock Styleheat: One-shot high-resolution editable talking face generation
  via pretrained stylegan.
\newblock In {\em ECCV}, 2022.

\bibitem{EgorZakharov2020FastBN}
Egor Zakharov, Aleksei Ivakhnenko, Aliaksandra Shysheya, and Victor Lempitsky.
\newblock Fast bi-layer neural synthesis of one-shot realistic head avatars.
\newblock In {\em ECCV}, 2020.

\bibitem{EgorZakharov2019FewShotAL}
Egor Zakharov, Aliaksandra Shysheya, Egor Burkov, and Victor Lempitsky.
\newblock Few-shot adversarial learning of realistic neural talking head
  models.
\newblock In {\em ICCV}, 2019.

\bibitem{zhang2021styleswin}
Bowen Zhang, Shuyang Gu, Bo Zhang, Jianmin Bao, Dong Chen, Fang Wen, Yong Wang,
  and Baining Guo.
\newblock Styleswin: Transformer-based gan for high-resolution image
  generation.
\newblock In {\em CVPR}, 2022.

\bibitem{RichardZhang2018LPIPS}
Richard Zhang, Phillip Isola, Alexei~A. Efros, Eli Shechtman, and Oliver Wang.
\newblock The unreasonable effectiveness of deep features as a perceptual
  metric.
\newblock In {\em CVPR}, 2018.

\bibitem{ZhimengZhang2022FlowguidedOT}
Zhimeng Zhang, Lincheng Li, and Yu Ding.
\newblock Flow-guided one-shot talking face generation with a high-resolution
  audio-visual dataset.
\newblock In {\em CVPR}, 2021.

\bibitem{RuiqiZhao2021SparseTD}
Ruiqi Zhao, Tianyi Wu, and Guodong Guo.
\newblock Sparse to dense motion transfer for face image animation.
\newblock In {\em ICCV}, 2021.

\end{thebibliography}
}

\clearpage
\begin{appendices}

\section{Implementation Details}
\label{sec:Implementation Details}
\noindent\textbf{Dataset.}
Following previous work~\cite{Drobyshev22MegaPortraits}, we train our warping and refinement networks on cropped VoxCeleb2 dataset~\cite{JoonSonChung2018VoxCeleb2DS}, which consists of 145k videos from 6k different identities. 
We preprocess the videos by cropping the faces with bounding boxes containing the landmarks from the first frame and resize each video sequence to $256 \times 256$ resolution.
We randomly select 500 videos from the VoxCeleb2 for evaluation. 
We use the source and driving frames from the same identity for training and same-identity reenactment evaluation, where the driving frames is also the ground truth image. For cross-identity reenactment evaluation, we randomly shuffle the identity in the previous test set, where the source and driving frames have different identities.

\noindent\textbf{Training details.}
We train the $256\times256$ base model on the VoxCeleb2 dataset with batch size of 48 using Adam optimizer of learning rate $2\times10^{-4}$ on $8\times$Tesla V100 GPUs. We set hyperparameters of losses as: $\lambda_{\textnormal{r}}=10, \lambda_{\textnormal{id}}=20, \lambda_{\textnormal{eye}}=50, \lambda_{\textnormal{mouth}}=50$ and $\lambda_{\textnormal{adv}}=1$.
We first train the warping network for 200,000 iterations, then the warping and refinement network jointly for 200,000 more iterations.

We further conduct our meta-learning stage for $N=14,000$ outer iterations and meta-learning rate $\beta=2\times10^{-5}$. In each iteration, we train an inner loop for $K=24$ iteration with inner loop learning rate $\alpha=2\times10^{-4}$ on
48 images per identity.

We train our temporal super-resolution module on the HDTF dataset for 20,000 iterations with batch size 8 and video sequence length 7 using Adam optimizer of learning rate $1\times10^{-4}$.

\noindent\textbf{Metrics.} 
Following previous works~\cite{ChenyangLei2020BlindVT,Lai2018LearningBVTC}, we evaluate our temporal consistency using warping error $E_{\text {warp }}$. For each frame $O_t$, we calculate the warping error with previous frame $O_{t-1}$ as:
\begin{eqnarray}
E_{\text {pair}}\left(t\right) = \frac{ \sum_{i=1}^N M_{t}\left(i\right)\left\|y_t(i)-W\left(y_{t-1}\right)(i)\right\|_1}{\sum_{i=1}^N M_{t}\left(i\right)}, \\
E_{\text {warp }}\left(\left\{t\right\}_{t=1}^T\right)=\frac{1}{T-1} \sum_{t=2}^T\left\{E_{\text{pair}}\left(t\right)\right\},
\end{eqnarray}
where $M_{t}$ is the occlusion map ~\cite{RuderDB2016VideoStyle} for a pair of images $y_t$ and $y_{t-1}, N$ is the number of pixels, and $W$ is backward warping operation with optical flow~\cite{ZacharyTeed2020RAFTRA}. The averages warping error $E_{\text {warp }}\left(\left\{t\right\}_{t=1}^T\right)$ is used to evaluate our temporal consistency.

\noindent\textbf{Approach to perform Cross ID reenactment.} We can reconstruct an accurate 3D face by fitting a morphable face model~\cite{JamesBooth2016A3M} based on the dense facial landmarks~\cite{wood2022dense}, which well disentangles identity with expression and motion. Benefiting from this, when performing challenging cross-identity reenactment, we simply combine the identity coefficients from the source 3D face with expression and head motion coefficients from the driving face and obtain a new 3D face. Then we project the resultant 3D face to 2D landmarks to serve as the driving target. In this way, there is no leakage of the driving identity so that the source identity could be well preserved.

\begin{figure}[t]
    \centering
    \begin{overpic}
        [width=1\columnwidth]{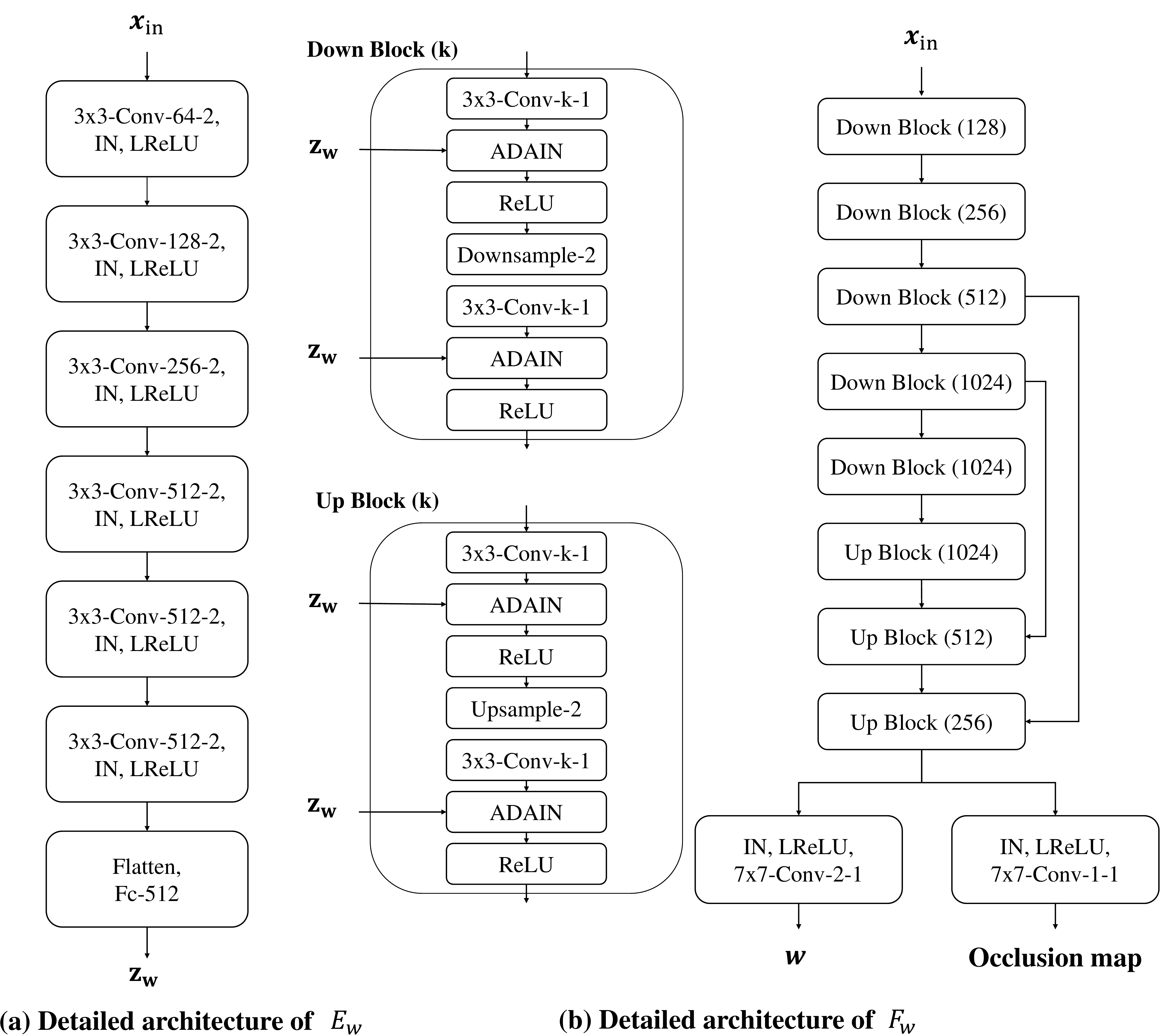}
    \end{overpic}
    \caption{Detailed architecture of our warping network.}  
    \label{fig:arch_warping}
\end{figure}

\begin{figure}[t]
    \centering
    \begin{overpic}
        [width=1\columnwidth]{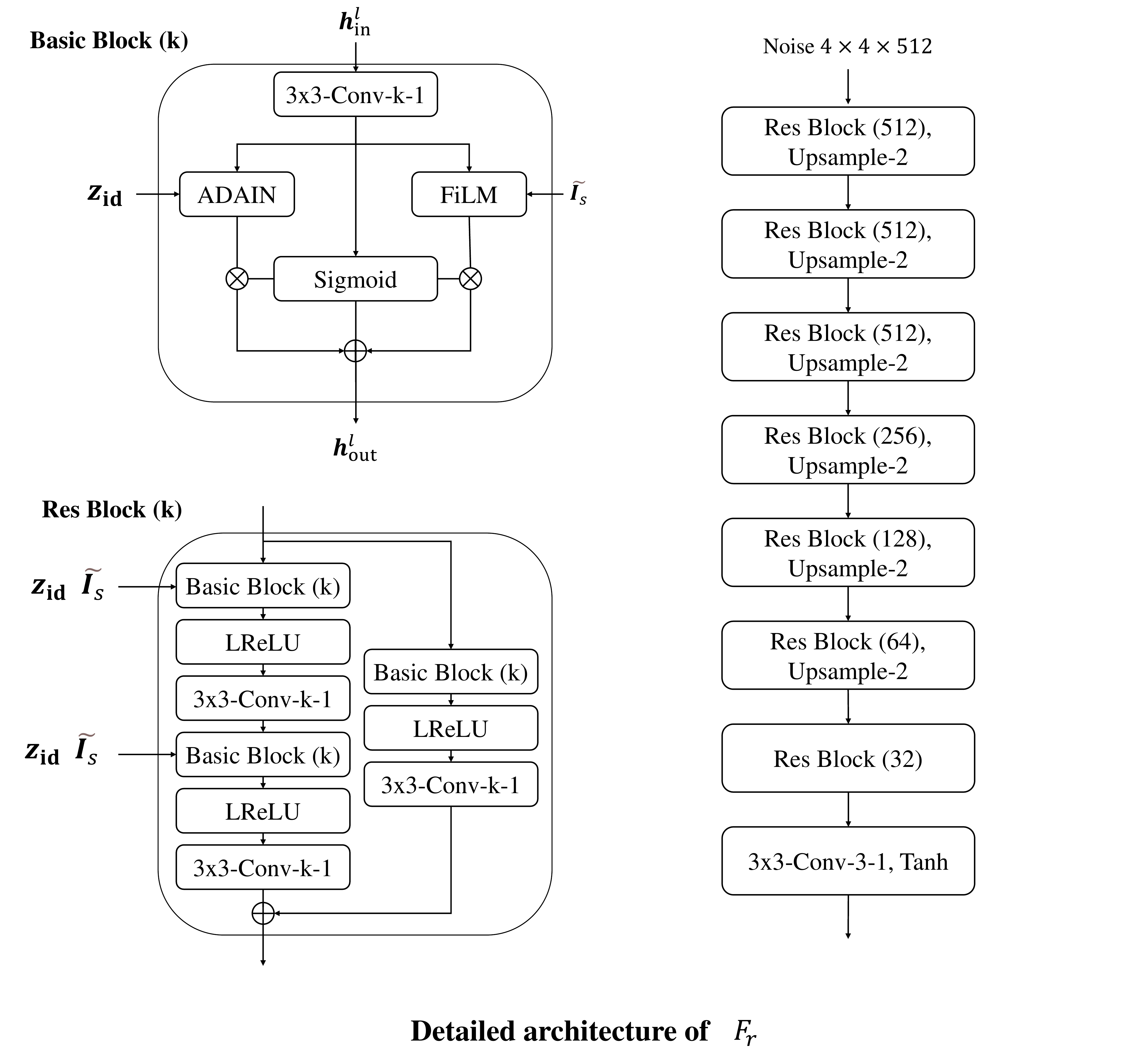}
    \end{overpic}
    \caption{Detailed architecture of our refinement network.}  
    \label{fig:arch_refinement}
\end{figure}

\begin{figure}[t]
    \centering
    \begin{overpic}
        [width=1\columnwidth]{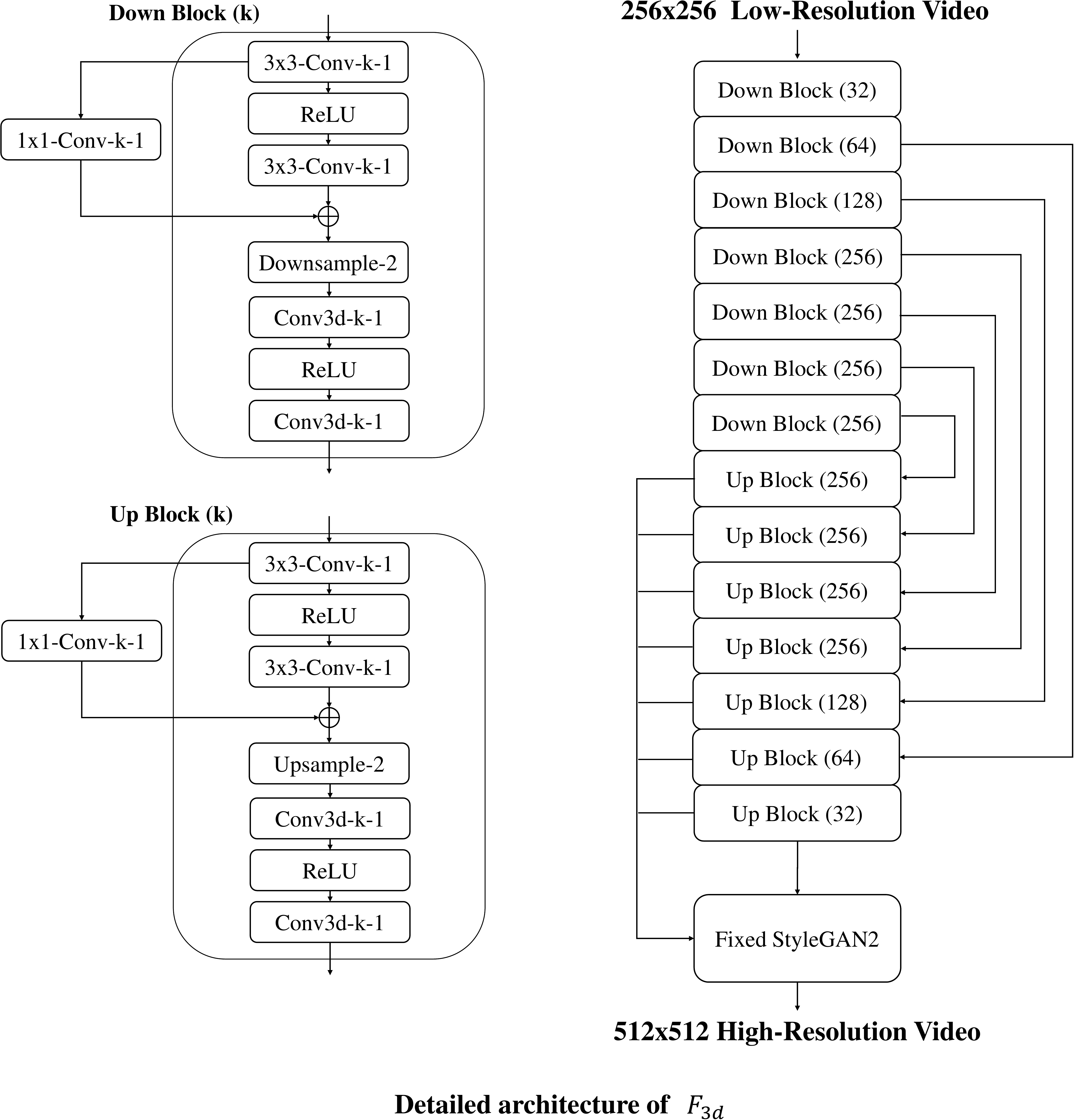}
    \end{overpic}
    \caption{Detailed architecture of our temporal super-resolution network.}  
    \label{fig:arch_tsr}
\end{figure}

\noindent\textbf{Detailed architecture.} The detailed architecture of our warping and refinement network is shown in Figure~\ref{fig:arch_warping} and Figure~\ref{fig:arch_refinement}. ``$3\times3\textnormal{-Conv-}k\textnormal{-}1$'' indicates a convolutional layer with kernel size of $3$, channel dimensions of $k$ and stride of $1$. ``LReLU, ReLU'' indicates LeakyReLU~\cite{https://doi.org/10.48550/arxiv.1505.00853} and ReLU~\cite{agarap2018deep} activation function respectively. 
Figure~\ref{fig:arch_tsr} illustrates the architecture of our temporal super-resolution network, where ``Conv3d-k-1'' represents a 3D convolution over temporal and spatial dimensions with $k$ feature dimensions and stride of $1$.

\section{Additional Ablation}
~\label{sec:Additional ablation}
\subsection{Visualization of ID, Landmark Ablation}

\begin{figure*}[t]
    \center
    \small
    \setlength\tabcolsep{1pt}
    {
    \renewcommand{\arraystretch}{0.6}
    \begin{tabular}{@{}cccc@{}}
         \includegraphics[width=0.35\columnwidth]{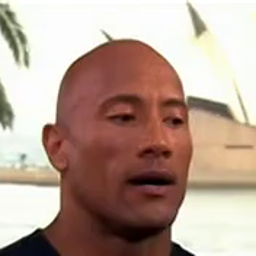} & \includegraphics[width=0.35\columnwidth]{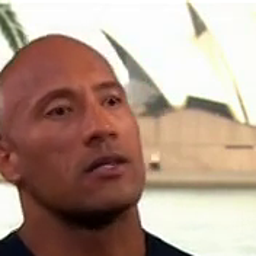} & \includegraphics[width=0.35\columnwidth]{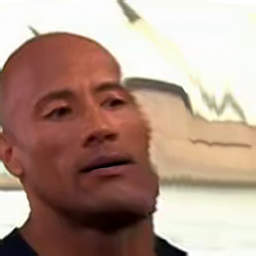} & \includegraphics[width=0.35\columnwidth]{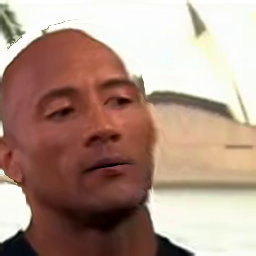}\\
         \includegraphics[width=0.35\columnwidth]{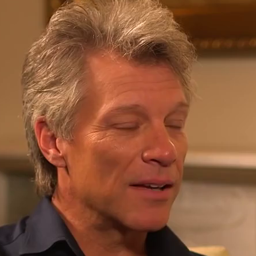} & \includegraphics[width=0.35\columnwidth]{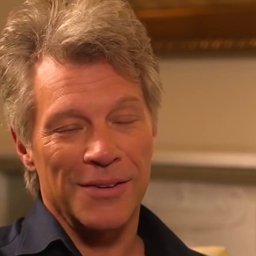} & \includegraphics[width=0.35\columnwidth]{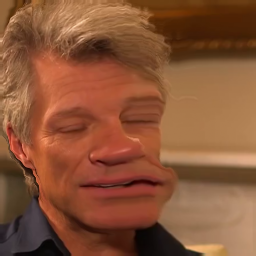} & \includegraphics[width=0.35\columnwidth]{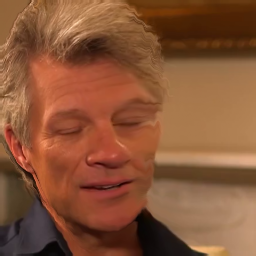}\\
         \includegraphics[width=0.35\columnwidth]{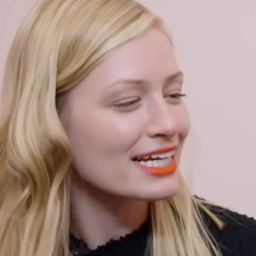} & \includegraphics[width=0.35\columnwidth]{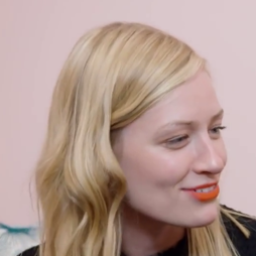} & \includegraphics[width=0.35\columnwidth]{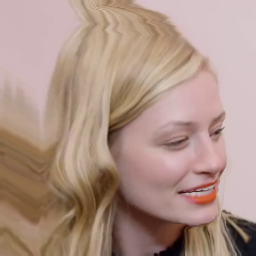} & \includegraphics[width=0.35\columnwidth]{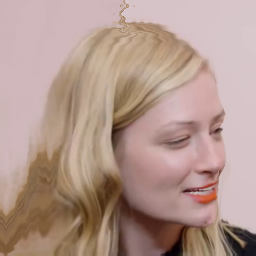}\\
          & & & \\
         Source & Driving & Sparse ldmks & Ours
    \end{tabular}
    }
    \caption{Qualitative comparison of warping quality of sparse landmarks and dense landmark encoding.}
    \label{fig:ablation_sparse}
\end{figure*}

\begin{figure*}[h]
    \center
    \small
    \setlength\tabcolsep{1pt}
    {
    \renewcommand{\arraystretch}{0.6}
    \begin{tabular}{@{}cccc@{}}
         \includegraphics[width=0.35\columnwidth]{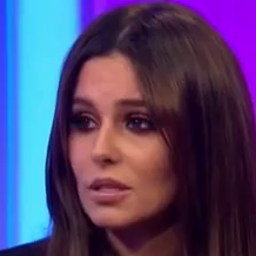} & \includegraphics[width=0.35\columnwidth]{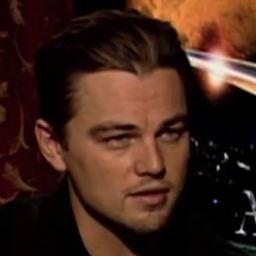} & \includegraphics[width=0.35\columnwidth]{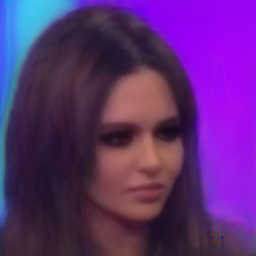} & \includegraphics[width=0.35\columnwidth]{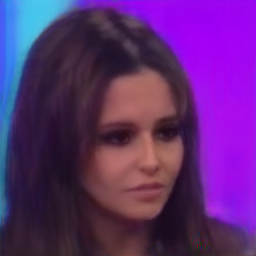}\\
         \includegraphics[width=0.35\columnwidth]{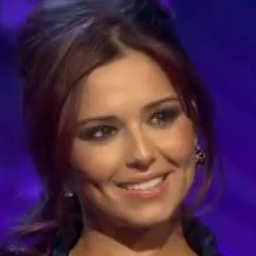} & \includegraphics[width=0.35\columnwidth]{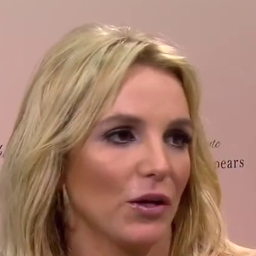} & \includegraphics[width=0.35\columnwidth]{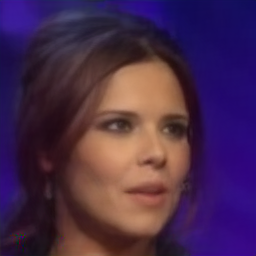} & \includegraphics[width=0.35\columnwidth]{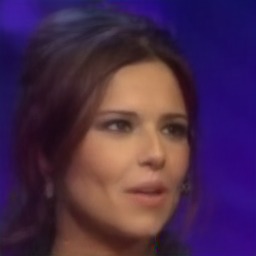}\\
         \includegraphics[width=0.35\columnwidth]{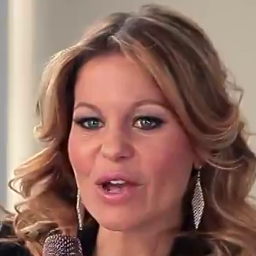} & \includegraphics[width=0.35\columnwidth]{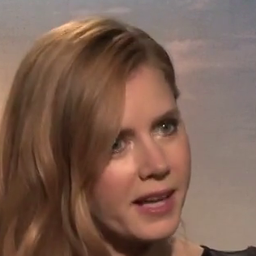} & \includegraphics[width=0.35\columnwidth]{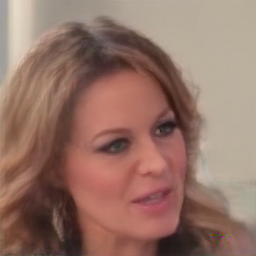} & \includegraphics[width=0.35\columnwidth]{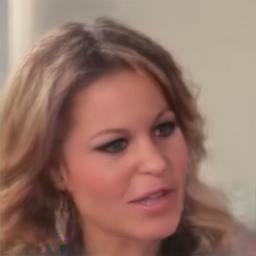}\\
          & & & \\
         Source & Driving & \makecell{Ours w/o ID \\ ID Loss 0.6272} & \makecell{Ours w/ ID \\ ID Loss 0.2112}
    \end{tabular}
    }
    \caption{Qualitative comparison of identity-preserving architecture.}
    \label{fig:ablation_ID}
\end{figure*}

Figure~\ref{fig:ablation_sparse} illustrates the warped images using the flow field produced by the warping network to evaluate the effectiveness of our dense landmark. 
The results guided by our dense landmarks are more accurate without obvious artifacts.
In Figure~\ref{fig:ablation_ID}, we show the visual changes brought by our ID-preserving refinement. Our source identity is better preserved, especially in the area of eye makeup and dimple.

\subsection{Ablation of Temporal Super-Resolution}
\begin{figure}[t]
    \center
    \footnotesize
    \setlength\tabcolsep{1pt}
    {
    \renewcommand{\arraystretch}{0.6}
    \begin{tabular}{@{}c}   
         \includegraphics[width=1.0\columnwidth]{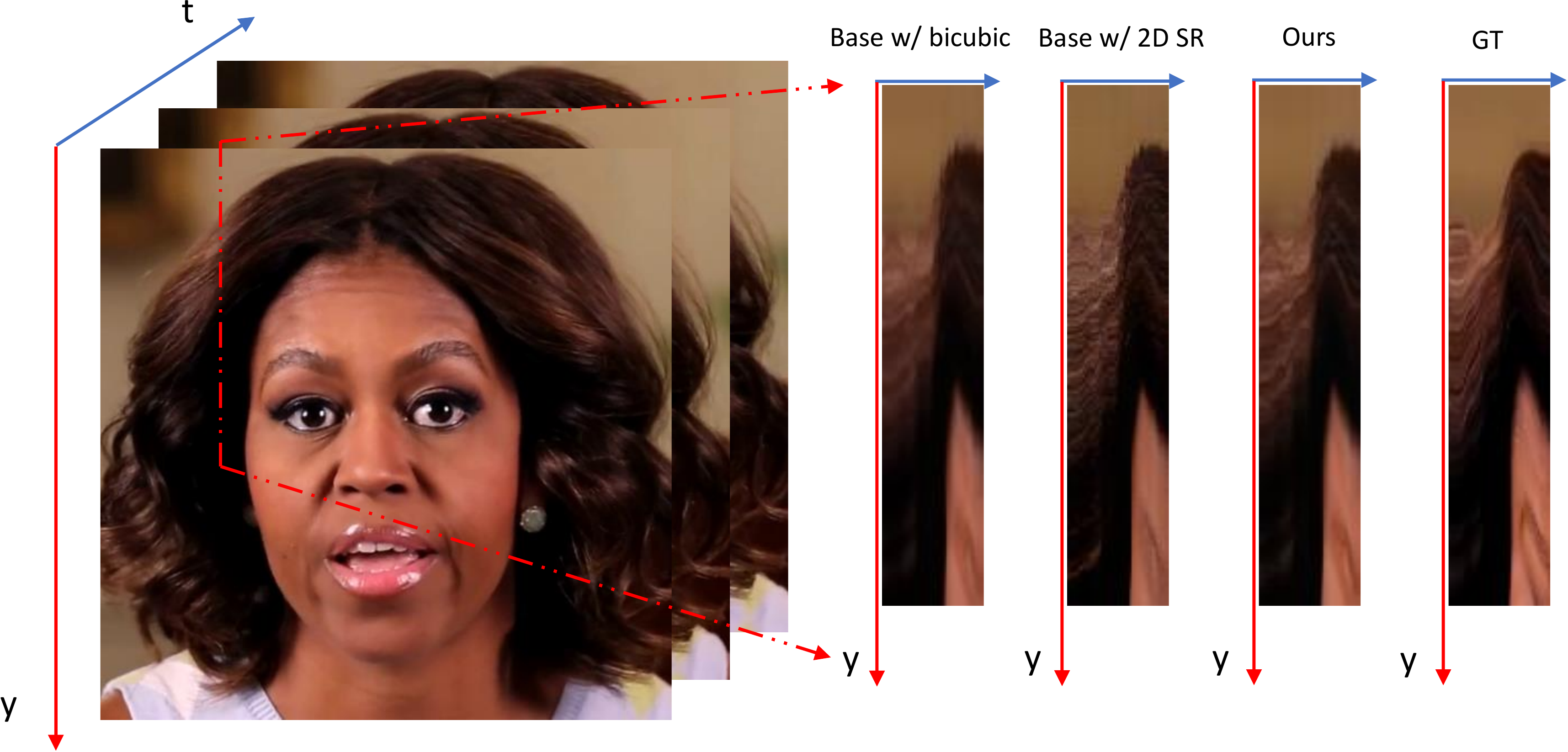}  \\
    \end{tabular}
    }
    \caption{Comparison of temporal profile. We select a column and observe its changes with time index. The result of our temporal super-resolution is more stable and consistent without flickering noise.} 
    \label{fig:temp_profile}
\end{figure}

In Figure~\ref{fig:temp_profile}, we select a column of the generated frame and visualize its temporal change. The bicubic-upsampled video lack texture of hair. The naive 2D face restoration baseline~\cite{wang2021gfpgan} generates more flickering artifacts. Our results have clear and stable temporal motion, which is close to ground truth.
In Table~\ref{table: temporal_consistency}, we provide an additional comparison with StyleHEAT~\cite{FeiYin2022StyleHEATOH} at $512\times512$ resolution and evaluate baselines using FVD~\cite{Unterthiner2019FVD}, which is a popular metrics in video generation. Our method achieves the lowest FVD, which demonstrates our high temporal fidelity. Since StyleHEAT generates unrealistic over-smooth and flickering images, its FID, LPIPS, and FVD are much worse than ours. Note that L1 loss $E_\textnormal{warp}$ is biased towards over-smoothed results. Thus, the $E_\textnormal{warp}$ of StyleHEAT and bicubic upsampled video can be lower than ours.

\begin{table}[t]
    \footnotesize
    \centering
    \begin{tabular}{l|c|c|c|c}
        \toprule
        {Methods} & FID$\downarrow$ & LPIPS$\downarrow$ & FVD$\downarrow$ & $E_{warp}\downarrow$ \\
        \midrule

        Ground Truth & - & - & -  & \textbf{0.0182}  \\
        \midrule
        StyleHEAT\cite{FeiYin2022StyleHEATOH} & 44.5207 & 0.2840 & 572.363 & 0.0207 \\
        \midrule
Base w/ bicubic & 25.5762 & 0.2285 & 248.413  & 0.0184  \\
        
        Base w/ GFPGAN~\cite{wang2021gfpgan} & 22.6351 & 0.2178 & 172.754  & 0.0242  \\

        \cellcolor{gray}\emph{Ours}
        &\cellcolor{gray}\textbf{21.4974}  & \cellcolor{gray}\textbf{0.2079} & \cellcolor{gray}\textbf{162.685} & \cellcolor{gray}0.0213   \\
        \bottomrule
    \end{tabular}
    \caption{Quantitative evaluation of our temporal super-resolution on self-reconstruction at $512\times512$ resolution.}
    \label{table: temporal_consistency}
\end{table}

\begin{table*}[t]
    \footnotesize
    \centering
    {
    \begin{tabular}{l|c|c|c|c|c|c|c|c}
        \toprule
        \multirow{2}*{Methods} & \multirow{2}*{Params} & \multirow{2}*{FPS} & \multicolumn{4}{|c|}{\textbf{Same-ID $256^2$}} & \multicolumn{2}{c}{\textbf{Cross-ID $256^2$}} \\
        \cmidrule{4-9}
        ~ & ~ & ~ & FID$\downarrow$ & FVD$\downarrow$ & LPIPS$\downarrow$ & ID Loss$\downarrow$ & FID$\downarrow$ & ID Loss$\downarrow$ \\
        \midrule
        FOMM~\cite{AliaksandrSiarohin2019FirstOM}     & 59.80  & 51.57 & 22.7112  & 136.4454           & 0.1577 & 0.0848 & 37.4306  & 0.4368\\
        PIRender~\cite{YuruiRen2021PIRendererCP} & 22.52  & 9.72  & 28.2376  & 367.3942          & 0.1881 & 0.1200 & 40.4600  & 0.3600\\
        DaGAN~\cite{hong2022depth}    & 60.36  & 29.04 & 21.0879  & \textbf{108.5139}  & 0.1427 & 0.0912 & 34.1784  & 0.4640\\
        DAM~\cite{tao2022structure}      & 59.75  & 43.74 & 23.7192  & 140.8459           & 0.1363 & 0.0832 & 40.4675  & 0.4400\\
        ROME~\cite{Khakhulin2022ROME}     & 123.85 & 2.63  & 119.9319 & 1204.52          & 0.5422 & 0.3376 & 102.9575 & 0.5360\\
        \cellcolor{gray}\emph{Ours} & \cellcolor{gray}130.28 & \cellcolor{gray}16.78 &\cellcolor{gray}\textbf{18.1581}  & \cellcolor{gray}219.6183 & \cellcolor{gray}\textbf{0.1335} & \cellcolor{gray}\textbf{0.0496} & \cellcolor{gray}\textbf{25.1646} & \cellcolor{gray}\textbf{0.1920} \\
        \midrule
        {Methods} & Params & FPS & \multicolumn{4}{|c}{\textbf{Same-ID $512^2$}} & \multicolumn{2}{|c}{\textbf{Cross-ID $512^2$}} \\
        \midrule
        StyleHEAT~\cite{FeiYin2022StyleHEATOH} & 367.70 & 0.03 & 41.3364 & 244.6287 & 0.2957 & 0.2560 & 136.3959 & 0.4960 \\
        \cellcolor{gray}\emph{Ours} & \cellcolor{gray} 284.97 & \cellcolor{gray}1.22 & \cellcolor{gray}\textbf{21.1314} & \cellcolor{gray}\textbf{131.8511} & \cellcolor{gray}\textbf{0.2150} & \cellcolor{gray}\textbf{0.0880} & \cellcolor{gray}\textbf{127.3204} & \cellcolor{gray}\textbf{0.2544} \\
        \bottomrule
    \end{tabular}
    }
    \caption{Evaluation against more baselines on a larger scale test set}
    \label{table: additional comparison}
    \vspace{-0.4cm}
\end{table*}

\begin{table}[t]
    \footnotesize
    \centering
    \begin{tabular}{l|c|c|c}
        \toprule
        {Methods} & Quality$\downarrow$ & Identity$\downarrow$ & Motion$\downarrow$ \\
        \midrule
        FOMM~\cite{AliaksandrSiarohin2019FirstOM} & 3.48 & 3.25 & 2.99 \\
        PIRender~\cite{YuruiRen2021PIRendererCP}  & 3.12 & 2.92 & 3.28 \\
        DaGAN~\cite{hong2022depth} & 3.48 & 3.55 & 3.01 \\
        DAM~\cite{tao2022structure} & 3.65 & 3.58 & 3.02 \\
        ROME~\cite{Khakhulin2022ROME} & 2.77 & 2.87 & 3.14 \\
        StyleHEAT~\cite{FeiYin2022StyleHEATOH} & 2.99 & 3.26 & 3.71 \\
        \cellcolor{gray}\emph{Ours}
        & \cellcolor{gray}\textbf{1.51} & \cellcolor{gray}\textbf{1.57} & \cellcolor{gray}\textbf{1.84}\\
        \bottomrule
    \end{tabular}
    \caption{Average ranking score of user study. User prefer ours the best in both three aspects.}
    \label{table: user study}
\end{table}

\section{Additional Comparison Results}
~\label{Sec:Additional Comparison Results}
\subsection{Additional Qualitative Comparison with Recent Methods on a Larger Scale Test Set}

\textbf{Main results.} We perform our evaluation including recent methods~\cite{hong2022depth,tao2022structure,Khakhulin2022ROME} on a larger test set, which contains 20 test videos following the setting of StyleHEAT~\cite{FeiYin2022StyleHEATOH}. For the same-id case, we evaluate using 500 frames of each video with 10k frames in total, while for the cross-id case, we use 1000 source images from CelebA-HQ as source images and use 100 frames of each video to drive 50 source images with 100k frames in total. The results are shown in Table~\ref{table: additional comparison}, in which our method achieves the best scores in almost all the metrics. Moreover, the FVD score of our full model improves significantly compared with the base model, which illustrates the effectiveness of the proposed temporal super-resolution network. Note that the compared methods target the one-shot setting and inevitably exhibit artifacts. In contrast, we are the first to study a personalized model which is of practical significance, and the proposed fast personalization is orthogonal to prior techniques and can be generally applied.

\textbf{Number of parameters and runtime (FPS).} We also provide the comparison of the model size and throughput between our model and other methods in Table~\ref{table: additional comparison}.

\textbf{User study.} We also conduct a user study to obtain the user's subject evaluation of different approaches. We present all the results produced by each comparing method to the participants and ask them to rank the score from 1 to 7 (1 is the best, 7 is the worst) on three perspectives independently: the image quality, the identity preservation and the motion drivability. 20 subjects are asked to rank different methods with 15 sets of comparisons in each study. The average ranking is shown in Table~\ref{table: user study}. Our method earns user preferences the best in both three aspects.

\subsection{Additional Qualitative Comparison}

We also provide additional videos on the \href{https://meta-portrait.github.io/}{webpage} to evaluate our results qualitatively. ``Ours-Base'' in the video denotes our base model in Sec 3.1, while ``Ours-Full'' denotes our full model with temporal-consistent super-resolution network. Our model is able to provide state-of-the-art generation quality with high temporal fidelity on both self-reconstruction and cross-reenactment tasks. Moreover, the videos of fast personalization illustrate the strong adaptation capability of our meta-learned model. The in-the-wild examples also demonstrate the generalized ability of the proposed model.

\begin{figure}[t]
    \center
    \footnotesize
    \setlength\tabcolsep{1pt}
    {
    \renewcommand{\arraystretch}{0.6}
    \begin{tabular}{@{}ccc@{}}
         \begin{overpic}
        [width=0.3\columnwidth]{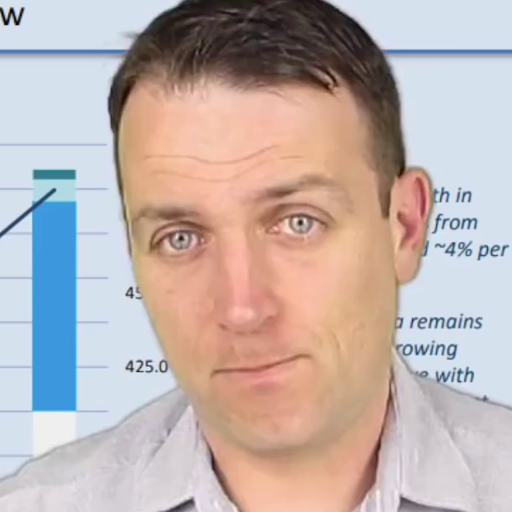}
        \end{overpic}
          & \begin{overpic}
        [width=0.3\columnwidth]{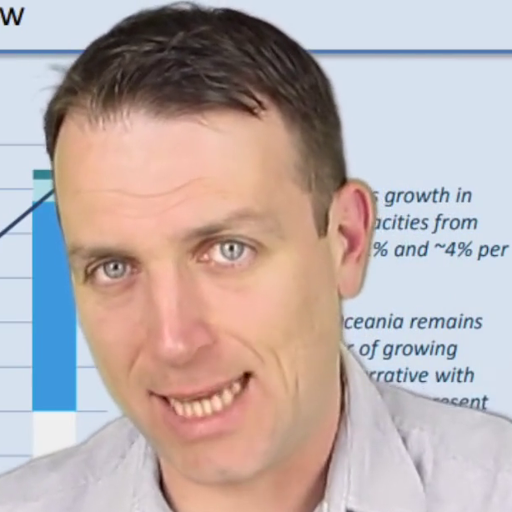}
        \end{overpic} 
        & \begin{overpic}
        [width=0.3\columnwidth]{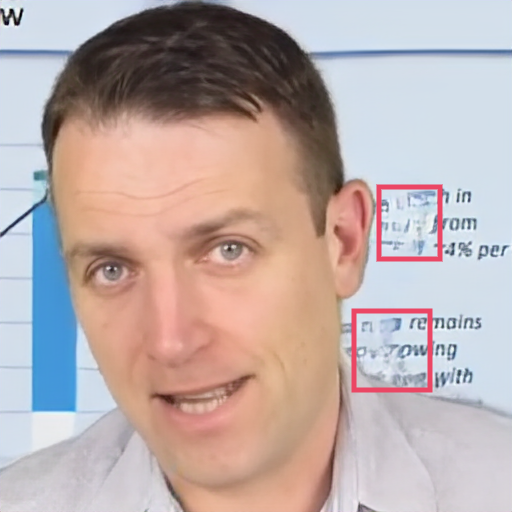}
        \end{overpic}\\
         Source & Driving & Ours
    \end{tabular}
    }
    \caption{Failure case of our framework.}
    \label{fig:failure_case}
\end{figure}

\section{Limitation}
~\label{Sec:Limitation}
Our one-shot model may not handle occlusions well. As shown in Figure~\ref{fig:failure_case}, the occluded text in the background appears blurry in the output result. One possible solution is to inpaint the background from pretrained matting and combined it with the generation results using alpha-blending following ~\cite{Drobyshev22MegaPortraits}, which we leave for future work.

\clearpage

\end{appendices}

\end{document}